%% file: main.tex
\documentclass[10pt]{article}
\usepackage{arxiv}
\input{macros.tex}
\usepackage[round, semicolon, authoryear]{natbib} 

\pgfplotsset{compat=1.18} 

\title{Conjugate Natural Selection}

\author{%
  Reilly Raab \\
  Computer Science and Engineering\\
  University of California, Santa Cruz\\
  Santa Cruz, CA 95064 \\
  \texttt{reilly@ucsc.edu} \\
  \AND
  Luca de Alfaro \\
  Computer Science and Engineering \\
  University of California, Santa Cruz \\
  Santa Cruz, CA 95064 \\
  \texttt{luca@ucsc.edu} \\
  \And
  Yang Liu \\
  Computer Science and Engineering\\
  University of California, Santa Cruz\\
  Santa Cruz, CA 95064 \\
  \texttt{yangliu@ucsc.edu}
}

\hypersetup{
pdftitle={Conjugate Natural Selection},
pdfauthor={Reilly Raab, Luca de Alfaro, Yang Liu},
pdfkeywords={
  evolution,
  natural selection,
  Bayesian inference,
  Fisher information,
  non-convex optimization,
  stochastic process,
  parameter estimation
}}

\begin{document}

\maketitle

\begin{abstract}
We prove that Fisher-Rao natural gradient descent (FR-NGD) optimally approximates the continuous time replicator equation (an essential model of evolutionary dynamics), and term this correspondence ``conjugate natural selection''. This correspondence promises alternative approaches for evolutionary computation over continuous or high-dimensional hypothesis spaces. As a special case, FR-NGD also provides the optimal approximation of continuous Bayesian inference when hypotheses compete on the basis of predicting actual observations. In this case, the method avoids the need to compute prior probabilities. We demonstrate our findings on a non-convex optimization problem and a system identification task for a stochastic process with time-varying parameters.
\end{abstract}

\keywords{
  evolution \and
  natural selection \and
  Bayesian inference \and
  Fisher information \and
  non-convex optimization \and
  stochastic process \and
  parameter estimation
}


\section{Introduction}

\newcommand{\pt}{\partial}

\renewcommand{\r}{r} 

\newcommand{\x}{x} 
\newcommand{\X}{X} 
\newcommand{\n}{\mathfrak{n}} 

\newcommand{\DD}{\mathscr{D}}
\newcommand{\smid}{\mkern-3mu\mid\mkern-3mu}

\newcommand{\ngd}{FR-NGD\xspace}
\newcommand{\cns}{CNS\xspace}

\renewcommand{\HH}{\mathcal{H}}
\newcommand{\h}{h}
\newcommand{\HHH}{\mathcal{P}}
\renewcommand{\H}{H} 
\newcommand{\MM}{\mathcal{M}} 

\newcommand{\F}{\mathcal{I}} 
\newcommand{\Err}{\mathcal{E}} 
\renewcommand{\L}{\mathscr{L}} 
\newcommand{\avgL}{\mkern4.5mu\overline{\mkern-4.5mu\L}\mkern-3mu} 

\newcommand{\avgu}{\overline{u}}

Evolution describes how distributions change. Specifically, evolution provides a
model for how a population's distribution of traits or strategies (hereafter
\emph{hypotheses}) changes over time as an environment modulates reproduction
rates (\ie, of individuals or of hypotheses; \citealp{sep-selection-units}):
Hypotheses that have higher \emph{fitness} are \qt{selected} by the environment
and, in expectation, become more popular with time. \textbf{The replicator
equation is a formal, analytic model of evolution} and is indispensable to
biology \citep{sinervo2006developmental, queller2017fundamental}.

In the replicator equation, the absolute fitness (in this paper, the negative
\emph{loss} \(\L\)) of hypotheses \(\h \in \HH\) is identified with its rate of
\emph{replication}: exponential growth (or decline) in a population where
different hypotheses compete for relative frequency \(\rho(\h) \in [0, 1]\).
For probability distributions over hypothesis space \(\HH\), this equation
induces \emph{replicator dynamics}, selecting hypotheses with lower than average
loss. In continuous time, the replicator equation i
\begin{equation*}
  \dv{}{t}\rho(\h) = \rho(\h) \Big[ \avgL_{\rho} - \L(\h) \Big],
  \quad\text{where}\quad
  \avgL_{\rho} \define \sum_{\h} \rho(\h) \L(\h), \quad \sum_{\h} \rho(\h) = 1.
\end{equation*}
The replicator equation has been applied to game theory
\citep{hofbauer1998evolutionary, sandholm2010population, cressman2014replicator,
friedman2016evolutionary}, economics \citep{friedman1991evolutionary}, and
machine learning \citep{hennes2019neural}.  For many real-world applications,
however, \textbf{an exceedingly large or continuous hypothesis space \(\HH\)
presents challenges} \citep[Sec~4.2]{bloembergen2015evolutionary}, and standard
techniques in evolutionary computation resort to modeling finite populations of
individuals directly \citep{back2018evolutionary}. Our results show promise in
alleviating these challenges. Specifically, we show that Fisher-Rao Natural
Gradient Descent (\ngd) optimally approximates replicator dynamics for a
tractable, lower dimensional representation of an evolving distribution over
\(\HH\).

Why should we want analytic model of evolutionary dynamics with large hypothesis
spaces? Suppose we wish to find the minimum value of a (possibly non-convex)
function \(\L \colon \RR^{d} \to \RR\), where we must make (possibly noisy,
expensive) queries for \(\L(\h)\) at any input \(\h \in \RR^{d}\). This problem
formulation is foundational to machine learning. First, we relax our attention
to \emph{individual} hypotheses \(\h\). Evolution describes how
\emph{distributions} change, and the replicator equation gives us a way to model
how any probability distribution \(\rho\) over hypothesis space
\(\HH = \RR^{d}\) should evolve when hypotheses are selected according to the
loss function \(\L\): Under replicator dynamics, hypotheses with minimal \(\L\)
will eventually out-compete all others, and any initial \(\rho\) that assigns
non-zero probability (density) to all of \(\HH\) will converge to a distribution
over globally optimal hypotheses. We call \(\rho\) a \emph{metahypothesis}.

For \emph{small} (finite) hypothesis spaces \(\HH\), the replicator dynamics may
be simulated directly, corresponding to the method of multiplicative weight
updates \citep{littlestone1994weighted, friedman2016evolutionary}. For
\emph{large} \(\HH\), however, our need to independently track \(\rho(\h)\) and
sample \(\L(\h)\) for each value of \(\h \in \HH\) causes issues: Note that the
replicator equation describes dynamics in \(\HHH\), where \(\HHH\) denotes the
space of probability distributions over \(\HH\). In general, the space
requirements for storing and manipulating arbitrary probability distributions
\(\rho \in \HHH\) grow proportional to the size of \(\HH\) (\ie, asymptotically,
as \(\Theta(|\HH|)\)). This is not feasible when \(\HH\) is continuous or
high-dimensional.

When a (meta)hypothesis space (\eg, \(\HHH\)) is large, it is standard in
multiple fields to choose a parametric manifold
\(\MM \define \{\rho(~\cdot~ ; \theta) \colon \theta \in \RR^{n}\} \subseteq \mathscr{\HHH}\)
of tractable solutions, using a parameter vector \(\theta \in \RR^{n}\), where
\(n\) dictates computational space requirements.  For example, we often
linearize dynamical systems near equilibrium using a manifold described by
eigenvalues and eigenvectors.  Similarly, neural networks parameterize manifolds
in function space using weights and biases.  Unfortunately, replicator dynamics
need not respect our chosen manifold: the replicator equation may force us off
of \(\MM\), demanding values of \(\rho\) that are not addressable by any
\(\theta\). To resolve this, we desire a model of evolutionary dynamics that is
closed on any parametric manifold \(\MM\) --- ideally one that approximates
replicator dynamics as closely as possible...

\textbf{Our central result is that Fisher-Rao Natural Gradient Descent (\ngd)
provides an optimal approximation of continuous replicator dynamics} constrained
to any twice-differentiable parametric manifold \(\MM\) (\cref{thm:cns}). We
refer to this correspondence as \emph{conjugate natural selection} (\cns;
\cref{sec:cns}). As an extension of this result, by building on known
connections between the replicator equation and Bayes's rule, we prove that \ngd
also provides an optimal approximation of Bayesian inference when loss is
identified with a Kullback-Leibler divergence between predictions and
observations (\cref{sec:bayes}). We demonstrate applications of these findings
to a non-convex optimization problem (\cref{sec:cns-applications}) and parameter
estimation of a stochastic process (\cref{sec:oci-applications}).

By calling attention to the special case of \ngd among metrics for natural
gradient descent, our work highlights beauty in the natural world and provides
immediate applications. First, our result indicates a provocative correspondence
between learning algorithms informed by information geometry and evolutionary
processes driven by natural selection. Second, our experiments indicate that
\cns provides an alternative approach to evolutionary computation for non-convex
optimization and may be used for Bayesian system identification and parameter
estimation.


\subsection{Related Work}

Prior work has discussed mutual connections between natural gradient descent,
replicator dynamics, and Bayesian inference, though even when cast as a
synthesis of these previous results, our results retain novelty. In particular,
we are aware of no prior work that explicitly identifies \ngd as the best
approximation of evolutionary dynamics nor Bayesian inference \emph{for all
twice-differentiable parameterization schemes \(\rho(\h ; \theta )\)}.
Additionally, we believe our specific construction of continuous Bayesian
inference is novel.

While exact correspondence between the replicator equation and \ngd has been
previously identified for tabular or Boltzmann-Gibbs parameterized distributions
(as the corresponding mirror-descent update) \citep{harper2009information,
harper2011escort, harper2020momentum, bloembergen2015evolutionary,
gao2017properties, hennes2019neural, otwinowski2020information,
chalub2021gradient}, this identity is limited to the case where \(\MM = \HHH\),
and we are aware of no prior work that extends this correspondence to
under-parameterized approximations \(\MM \subset \HHH\).

While \citet{harper2009replicator, harper2009information} recognizes the
replicator equation as both an instance of \ngd and as an inference dynamic
guided by Fisher information geometry and connected to Bayes's rule in discrete
time, the cited work stops short of identifying the continuous time replicator
equation as a \emph{generalization} of Bayes's rule to continuous time, nor does
it identify average fitness with a (negative) Kullback-Leibler divergence
(despite recognizing the latter as a Lyapunov function for the replicator
equation).  Achieving deeper connections to Bayesian inference, recent work by
\citet{khan2021bayesian} shows that \ngd gives rise to optimal Bayesian
inference even for underparameterized distributions, but the provided analysis
assumes exponential families, rather than arbitrary, twice-differentiable
parameterizations.

The previous results cited above all generalize to arbitrary
twice-differentiable parameterizations in light of recent work by
\citet{nurbekyan2022efficient}, who observe that natural gradient descent with
any metric \(g(\theta)\) yields the least-squares optimal approximation in
\(\MM\) to natural gradient descent with metric \(g(\rho)\) on \(\HHH\)
\citep[Eq.  2.2]{nurbekyan2022efficient}. For our purposes: natural gradient
descent on a parametric manifold always yields an optimal approximation of
natural gradient descent in the continuous analog of the tabular setting (\ie,
in the case of \ngd, the replicator equation, where, for every \(\h\),
\(\rho(\h)\) may be independently specified). Nonetheless, to our knowledge,
this observation has not been synthesized with the aforementioned results, nor
have the implications for evolutionary dynamics been previously explored.


\section{Preliminaries} \label{sec:preliminaries}

Before detailing our results, we first review necessary background, establishing
our setting in \cref{sec:setting}. We briefly discuss properties of the
replicator equation in \cref{sec:evolutionary-dynamics} and provide essential
results from information geometry in \cref{sec:ngd}.

\subsection{Setting} \label{sec:setting}

In this paper, we denote a \emph{hypothesis} as \(\h\), which we identify with a
\qt{strategy} in the evolutionary game theory literature
\citep{friedman2016evolutionary}: For example, \(\h\) may represent a
combination of genes, a behavior, a belief, or a machine learning policy. Let
\(\HH\) denote the space of possible hypotheses, such that \(h \in \HH\) for
all \(h\). For example, \(\HH\) might represent a population's genome, a set of
competing social norms, an array of alternative beliefs, or the parameter space
of a neural network.  Finally, let \(\HHH\) denote the simplex, or space
of probability distributions, over \(\HH\).

\begin{table}[h]
  \caption{Variables} \label{table:notation}
  \centering
  \begin{tabular}{cl}
    \toprule

    $\h$ & a hypothesis. $\h \in \HH$. \\
    $\HH$ & hypothesis space (arbitrary in size and dimension). \\
    $\HHH$ & the space of probability distributions over \(\HH\). \\
    $\rho$ & a probability distribution over hypotheses. $\rho \in \HHH$. \\
    $\H$ & a random hypothesis. $\H \sim \rho$. \\
    $\L$ & a (possibly noisy) loss function. \(\L \colon \HH \to \RR\). \\
    $\avgL_{\rho}$ & the expected value of \(\L\) according to distribution \(\rho\). \\
    \midrule
    $\theta$ & a parameter value. $\theta \in \RR^{n}$. Components indexed, \eg, as \(\theta^{i}\). \\
    $\rho(\h ; \theta)$ & the probability (density) at \(\h\), parameterized by \(\theta\). \\
    $\MM$ & the parametric manifold $\{\rho(~\cdot~; \theta) \colon \theta \in \RR^{n}\}$.  \(\MM \subseteq \HHH\). \\
    $\F$ & the Fisher (\cref{sec:Fisher}). \\
    $s$ & score (\cref{defi:score}), \eg, \(s_{i}(\theta ; h)\). \\
    $\Err$ & ``natural deviation'' (\cref{defi:err}). \\

    \bottomrule
  \end{tabular}
\end{table}

We use \(\rho \in \HHH\) to represent an individual probability
distribution over \(\HH\), and denote a hypothesis sampled at random from this
distribution as \(\H \sim \rho\). In a slight abuse of notation, we denote the
probability (density) associated with \(\h\) in \(\rho\) as
\(\rho(h)\). When \(\HH\) is discrete, \(\rho(\h)\) corresponds to the relative
frequency of \(\h\) in a given population. When \(\HH\) is continuous,
\(\rho(\h)\) generalizes to a probability density, while sums over \(\h\)
generalize to integrals.  While all equations in this paper readily generalize
to continuous \(\HH\), we write our equations as if \(\h\) were discrete
for consistency and convenience. This is not a limitation of our results.

For ease of notation, let a dot above a symbol to denote its \emph{full} time
derivative (that is,
\(\forall u, \dot{u} \equiv \dd{u}\mkern-3mu/\mkern-3mu\dd{t}\)) and a bar over
a variable to denote its expectation value (explicitly,
\(\forall u, \avg{u} \equiv \E[u]\)). We denote contravariant vectors with an
upper index and covariant vectors with a lower index, using the Einstein
summation convention of implicitly summing over matching upper and lower indices
in a single term (formally,
\(\forall u, v, u_{i} v^{i} \equiv \sum_{i} u_{i} v^{i}\)), but we will not use
this convention for time index $t$.  We also use standard shorthands for partial
derivatives, identifying
\(
\pt_{i} (\cdot) \equiv {\pt(\cdot)/\pt \theta^{i}}
\)
and
\(
\pt_{t} (\cdot) \equiv {\pt(\cdot)/\pt t}
\).

The \textbf{motivating problem} we consider is the minimization, over \(\rho\), of
expected \emph{loss}, for some loss function \(\L \colon \HH \to \RR\), when
\(\H \sim \rho\). Put simply, we wish to select the distribution of hypotheses
\(\rho^{\star}\) with the smallest average loss.
\begin{equation} \label[problem]{eq:optim}
  \rho^{\star} = \argmin_{\rho}  \avgL_{\rho}
  \quad ; \quad
  \avgL_{\rho} \define
  \E_{H \sim \rho}\Big[ \L(H) \Big] =
  \sum_{\h} \rho(\h) \L(\h).
\end{equation}
In general, we assume that \(\L\) may be a non-convex function. For example,
\(\L(\h)\) could represent the rate of excess deaths compared to births for a
genotype \(h\), the negative rate of total returns for an investment portfolio
\(h\), or the expected loss of machine learning policy \(h\) on a given task.
For now, we assume that we may sample \(\L\) without noise, but this
condition is easily relaxed as long as noise remains unbiased.

Ultimately, our proposed solution for \cref{eq:optim} involves a
\textbf{twice-differentiable parametric manifold} \(\MM \subset \HHH\) of
distributions \(\rho(h ; \theta) \in \MM\) where \(\theta \in \RR^{n}\)
is a parameter vector for integer \(n\) greater than zero.  We analyze
continuous time equations of motion for \(\rho\) and \(\theta\): the replicator
equation for \(\rho\) and \ngd for \(\theta\). In \cref{sec:cns}, we show that
the latter optimally approximates the former.


\subsection{Replicator Dynamics} \label{sec:evolutionary-dynamics}

The replicator equation describes \emph{replicator dynamics}. As background, we
restate the continuous time replicator equation and introduce Price equation
(\cref{lem:price}). While our treatment throughout this paper assumes a
continuous time variable \(t\), we also derive the discrete-time form of the
replicator equation in the supplementary material
(\cref{lem:replicator-discrete}), taking care to explicitly consider time
intervals of the form \([t, t+\Delta t)\).

We have already introduced the continuous time replicator equation in a form
adapted to the notation of machine learning literature:
\begin{defi}\label{defi:replicator}
  \emph{The replicator dynamics} are governed by the equation
\begin{equation} \label{eq:replicator}
  \dot{\rho}(\h) = \rho(\h) \Big[ \avgL_{\rho} - \L(\h) \Big],
  \quad\text{where}\quad
  \avgL_{\rho} \define \sum_{\h} \rho(\h) \L(\h), \quad \sum_{\h} \rho(\h) = 1.
\end{equation}
\end{defi}
Although we allow ourselves to omit time-indexing for \(\rho\) and \(\L\), these
quantities are time-varying.
\begin{rem}
  \label{rem:no-mutation}
  \emph{(No new hypotheses)}.
For any finite times \(t\) and \(t'\),
\(
  \rho_{t}(\h) = 0
\)
iff
\(
  \rho_{t'}(\h) = 0
\).
\end{rem}
\begin{proof}
  The replicator equation has solutions of the form
  \(
  \rho_{t}(\h) = \rho_{0}(\h) \exp \int_{0}^{t} \left( \avgL_{\rho_{t'}} - \L_{t'}(\h) \right) \dd{t'}
  \),
  which does not admit roots in finite time unless \(\rho_{0}\) (and therefore
\(\rho_{t}\), for all \(t\)), is zero.
\end{proof}

\cref{rem:no-mutation} reveals that \textbf{the replicator equation cannot
generate new hypotheses}. For this reason, it is often combined with mutation or
diffusion terms in practice, but the resulting dynamics are  more difficult to
solve \citep{bloembergen2015evolutionary}. As an approximation of replicator
dynamics with under parameterized \(\rho(\theta)\), \textbf{\ngd can avoid this
issue}, since eliminated hypotheses can be reintroduced or provably never be
eliminated (\eg, when \(\rho(\h ; \theta)\) is everywhere non-zero by design).

\begin{restatable}[]{lem}{lemprice}\label{lem:price}
\emph{(The Price Equation)}. For any function or real-valued property of
hypotheses \(u \colon \HH \to \RR\), the expected value of \(u\), denoted \(\avgu_{\rho}\) when \(\h\) is
sampled with probability \(\rho(\h)\), evolves according to
\begin{equation} \label{eq:price}
  \dv{}{t} \avgu_{\rho}
  = -\Cov_{\H\sim\rho}\Big[u(\H), \L(\H)\Big]
  + \E_{\H\sim\rho}\Big[ \dot{u}(\H) \Big]
  \quad ; \quad
  \avgu_{\rho} \define  \E_{\H \sim \rho}\Big[u(\H)\Big].
\end{equation}
\end{restatable}

Many key results of evolutionary dynamics, such as fundamental theorems for gene
and phenotype selection or heritability, may be derived from the Price equation
\citep{queller2017fundamental}. We provide a derivation in the supplementary
material (\cref{prf:lemprice}), though this is a standard result.

%

\subsection{Fisher-Rao Natural Gradient Descent} \label{sec:ngd}

Variations of gradient descent have recently become standard techniques for
non-convex optimization problems like \cref{eq:optim}, facilitated by automatic
differentiation and parallelized updates \citep{fradkov2020early, tappertfrank}.
In broad strokes, the technique is to first differentiably parameterize a search
space \(\MM\) with a mapping \(\theta \mapsto \rho(~\cdot~ ; \theta)\), for
example, then repeatedly update \(\theta\) in a direction that approximates the
\qt{fastest} decreasing value (\ie, the negative gradient) of the expected loss
\(\avgL\).
\begin{defi}\label{defi:gradient-descent}
  \emph{Naive Gradient Descent}, in continuous time, is given by the update rule
  \begin{equation}\label{eq:gradient-decent}
    \dot{\theta}^{i} = - \pt_{i} \avgL.
  \end{equation}
\end{defi}
There is a problem with this update that is often unacknowledged in machine
learning pedagogy: One side of this equation is \emph{contravariant}, while the
other is \emph{covariant}. To understand this intuitively, assign units to the
quantities such that
\(\dim(\theta) = \mathsf{U}\), \(\dim(\L) = \mathsf{L}\), and
\(\dim(t) = \mathsf{T}\). It follows that
\(\dim(\dot{\theta^{i}})=\mathsf{UT^{-1}}\) while
\(\dim(\pt_{i}\avgL) = \mathsf{LU}^{-1}\). While a learning rate provides a
natural conversion between \(\mathsf{L}\) and \(\mathsf{T}^{-1}\), the powers of
\(\mathsf{U}\) do not balance on each side of \cref{eq:gradient-decent}, and the
equation is dimensionally invalid. This problem is resolved by explicit
consideration of an (inverse) metric \(g^{ij}\) with units \(\mathsf{U}^{2}\)
that assigns distances and angles in the cotangent space of \(\theta\) (\ie,
where \(\pt_{i}(\cdot), \pt_{j}(\cdot)\), \etc live). The metric \(g_{ij}\)
applies to the tangent space of \(\theta\) (\ie, where \(\dd{\theta}^{i}\),
\(\dd{\theta}^{j}\), \etc live).
\begin{defi}\label{defi:covariant-gradient-descent}
  \emph{Covariant Gradient Descent} is given by the update rule
  \begin{equation}\label{eq:covariant-gradient-decent}
    \dot{\theta}^{i} = - g^{ij} \pt_{j} \avgL.
  \end{equation}
\end{defi}
Importantly, \textbf{the choice of metric \(g\) can strongly influence the dynamics of
gradient descent}, which we call the gradient \emph{flow}. That is, the direction
of the \qt{fastest} decreasing value of \(\avgL\) depends on how the tangent
space of \(\theta\) is measured by \(g\). The implicit assumption of naive
gradient descent is that \(g^{ij} = \delta^{ij}\) for Kronecker delta \(\delta\)
with the appropriate units, \ie, a Euclidean metric for the (co)tangent space of
\(\theta\). \ngd, that is, natural gradient descent with respect to the
Fisher-Rao metric, uses a specific, alternative choice of \(g\) in
\cref{eq:covariant-gradient-decent} that derives from information geometry
\citep{amari1998natural, martens2020new}. It (un)warps the space around any
given parameter value \(\theta\) before performing the gradient update, so that
small updates of \(\theta\) in any direction all contain the same marginal
information about the new distribution \(\rho(\theta + \dd{\theta})\).  The
metric it uses is known as the \emph{Fisher}.

\begin{figure}[ht]
  \centering
    \centering
    \adjustbox{scale=0.75}{\includegraphics[trim=0.01cm 0cm 0cm 0cm, clip]{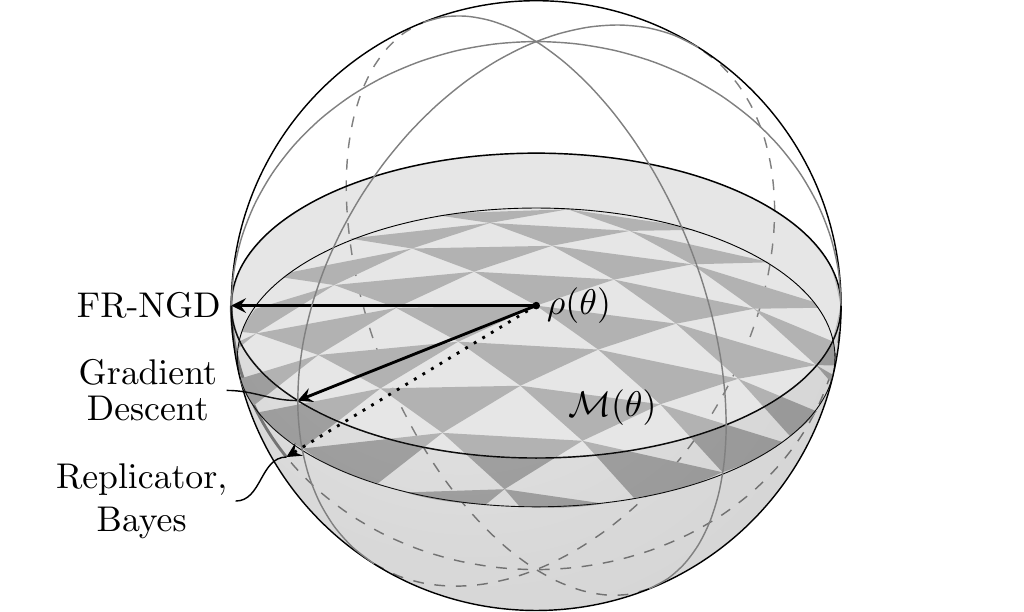}}
    \adjustbox{scale=0.75}{\includegraphics[trim=0.01cm 0cm 0.59cm 0cm, clip]{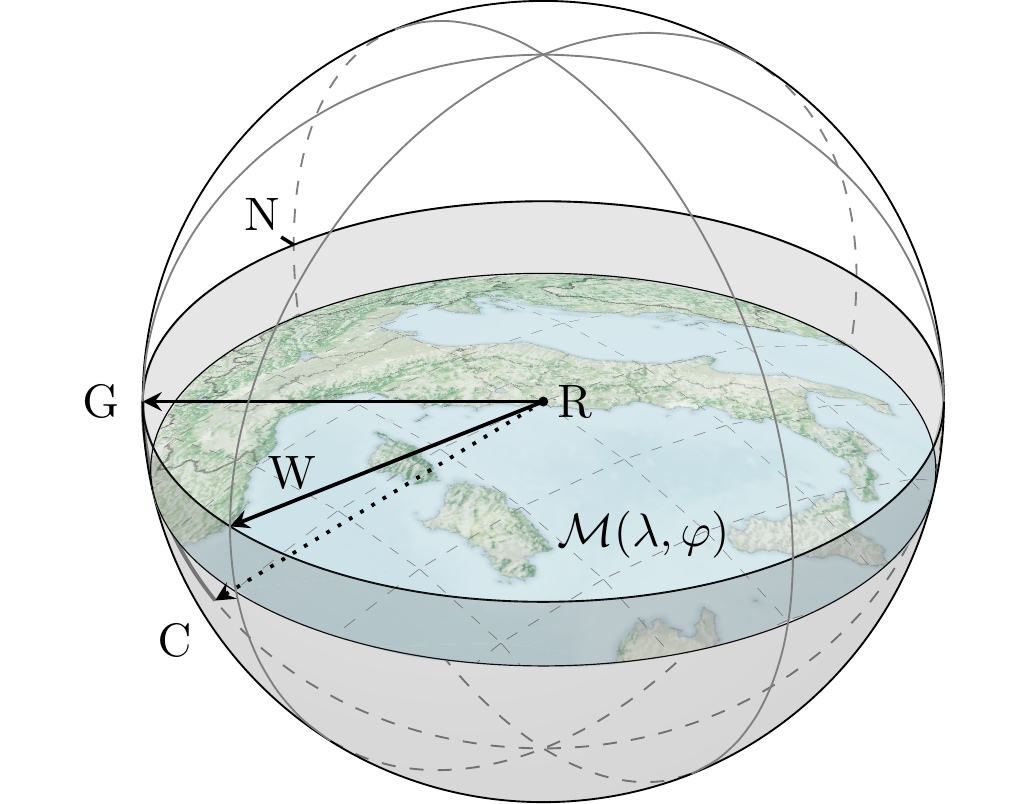}}
    \caption{An analogy for how different metrics can suggest different
parameter updates in \cref{eq:covariant-gradient-decent}. Earth's curvature is
exaggerated to emphasize that vectors W and G are tangent to its surface. The
direction of travel from Rome (point R) that most rapidly decreases one's
distance from Chicago, when measured in the Euclidean space of
latitude-longitude pairs \((\lambda, \varphi)\), is nearly due west (vector W),
because Rome and Chicago have nearly the same latitude \(\lambda\). Performing
gradient descent with an implicit Euclidean metric for parameter space is
similarly naive. Vector C is tangent to the true shortest path in physical
space: north-west at an angle of nearly 35 degrees downwards.  Like the update
given by the replicator equation or a continuous generalization of Bayes's rule,
this direction may not be tangent to the manifold \(\MM\). Constrained
to \(\MM\), the optimal approximation of the direction of C is its
projection, G: north-west, tangent to the surface, and tangent to the geodesic
from Rome to Chicago on the surface of the sphere. Map Data Credit: NASA Visible
Earth.}
\label{fig:earth}
\end{figure}

In \cref{fig:earth}, we give an analogy for how the choice of metric can affect
covariant gradient decent. Both a flat map and a globe ``warp'' our perception
of local distances and angles, and can mislead us when finding the ``fastest''
route between two points on Earth.

\subsubsection{The Fisher} \label{sec:Fisher}
The Fisher-Rao information metric tensor, Fisher information matrix (FIM), or
``Fisher'' may be expressed in multiple ways, but is defined thus:

\begin{restatable}{defi}{defifisher}\label{defi:fisher}
  \emph{The Fisher} for \(\theta\) is
\(
  \F_{ij}(\theta)
  \define \Cov_{H \sim \rho}\left[ \pt_{i} \log \rho(H ; \theta),~ \pt_{j} \log \rho(H ; \theta) \right].
\)
\end{restatable}
As a covariance matrix, \(\F\) is symmetric and positive semi-definite.  \(\F\)
fails to be fully positive definite (\ie, is degenerate) when parameter updates
in different directions (up to constant scaling) affect \(\rho\) identically.
Note that the quantity \(\pt_{i} \log \rho(\h ; \theta)\) appearing in
\cref{defi:fisher} is called the \emph{score}.
\begin{defi}
  \label{defi:score}
  \emph{The score} \(s_{i}(\theta ; \h)\) is defined as \(\pt_{i} \log \rho(\h ; \theta)\).
\end{defi}
\begin{lem}
  \label{lem:score}
  \emph{(Zero Expected Score)}. The expected score is zero. That is,
\(
  \E_{\H \sim \rho}\left[ \pt_{i} \log \rho(\H) \right] = 0.
\)
\end{lem}
\begin{proof}
  \(
    \quad
    \forall i, ~~
    \E_{\H \sim \rho}\left[ \pt_{i} \log \rho(\H) \right]
    = \sum_{\h} \rho(\h) \pt_{i} \log \rho(\h)
    = \sum_{\h} \pt_{i} \rho(\h)
    = \pt_{i} \sum_{\h} \rho(\h)
    = 0.
    \)
\end{proof}

\subsubsection{Primal Gradient Flow}

\begin{defi}
  \label{defi:ngd}
  \emph{The primal gradient flow}, for \(\theta\), induced by \ngd of \(\avgL\)
with respect to \(\theta\) is
\begin{equation} \label{eq:ngd}
  \dot{\theta}^{i} = - \big[\F^{\dagger}(\theta)\big]^{ij}\pt_{j} \avgL,
\end{equation}
\end{defi}
where \(\F\) is the Fisher and \(\F^{\dagger}\) is its Moore-Penrose inverse. While
\cref{eq:ngd} is often used in practice to update \(\theta\), the dynamics of
the \emph{distribution} \(\rho\) are of ultimate, material consequence.  The
dynamics of \(\rho\) are described by the conjugate gradient flow.

\subsubsection{Conjugate Gradient Flow}
\begin{restatable}{defi}{deficonjngd}\label{defi:conj-ngd}
  \emph{The conjugate gradient flow}, for \(\rho \in \MM\), induced by
\ngd of \(\avgL\) with respect to \(\theta\) is
\begin{equation} \label{eq:conj-ngd}
  \F_{ij}(\theta) \dot{\theta}^{j} = - \pt_{i} \avgL.
\end{equation}
\end{restatable}
\(\F(\theta)\) is positive definite and invertible  when \(\theta\) has only
non-degenerate degrees of freedom, in which case
\(\F^{\dagger}(\theta) = \F^{-1}(\theta)\), the nullspace of \(\F^{\dagger}(\theta)\) is
orthogonal to the tangent space of \(\MM\), and \cref{eq:conj-ngd} and
\cref{eq:ngd} are equivalent.
When \(\F(\theta)\) is not invertible, which occurs when \(\theta\) has
redundant degrees of freedom (\eg, in a state of gimbal lock), the properties of
the Moore-Penrose inverse imply that \cref{eq:ngd} solves \cref{eq:conj-ngd},
producing, among under-determined solutions for \(\dot{\theta}\), the one with
minimal Euclidean norm. In this case, the \emph{conjugate} gradient flow induced
by \ngd with respect to \(\theta\) is still described by \cref{eq:conj-ngd}.  We
show that this conjugate gradient flow is an optimal approximation of the
replicator equation (\cref{thm:cns}).

To prime intuition, using results provided in the supplementary material
(\cref{lem:conj-flow} and \cref{lem:conj-loss}), we may rewrite the conjugate
gradient flow (\cref{defi:conj-ngd}) as
\begin{equation}\label{eq:proj}
  \sum_{\h}  \Big[\underbrace{\pt_{i} \log \rho(h)}_{s_{i}}\Big] \dot{\rho}(\h)
  = \sum_{\h} \Big[ \underbrace{\pt_{i} \log \rho(\h)}_{s_{i}}\Big] \rho(\h) \Big(\avgL -  \L(\h) \Big).
\end{equation}
Comparing \cref{eq:replicator} and \cref{eq:proj}, we \textbf{recognize
conjugate gradient flow as a projection of the replicator dynamics} onto the
dual space of \(\theta\) spanned by the score (\cref{defi:score}), where the
score provides a basis for the tangent space of \(\MM\) at \(\theta\) such that,
by \cref{lem:score}, local motion along each basis vector introduces zero
marginal entropy relative to \(\rho(~\cdot~; \theta)\):
\begin{equation}
  \E_{\H\sim\rho}\Big[\log\rho(H) + s_{i}(\H) \dd{\theta}^{i}\Big] =
 \E_{\H\sim\rho}\Big[\log\rho(H)\Big]
\end{equation}
As a projection of replicator dynamics onto \(\MM\), we naturally expect the
conjugate gradient flow to \textbf{minimize an appropriately defined distance} from the
replicator dynamics.  Indeed, we define such a distance with \cref{defi:err} and
realize this expectation with \cref{thm:cns}.


\section{Conjugate Natural Selection} \label{sec:cns}

In this section, we state our primary results. We make use of the
Fisher metric for \(\rho\), also known as the Shahshahani metric
\citep{harper2009information}, which follows from \cref{defi:fisher} when
\(\theta \equiv \rho\):
\begin{equation} \label{eq:shahshahani}
  \F_{ij}(\rho) = \E_{\H\sim\rho}\left[\frac{\delta_{ij}}{\rho(H)^{2}}\right].
\end{equation}

\begin{restatable}[]{defi}{defierr}\label{defi:err}
  \emph{The natural deviation} \(\Err\) of \(\dot{\rho}\), induced by
\(\dot{\theta}\), from its nominal value under the replicator equation is given
by the corresponding mean-squared error in realized relative fitness
\(\dd/\mkern-4mu\dd{t} \log \rho\).
\begin{equation} \label{eq:err}
  \begin{aligned}
  \Err\big(\dot{\theta}\big) \define
  \frac{1}{2}
  \E_{\H \sim \rho}\left[
    \bigg( \underbrace{\dv{}{t} \log \rho(\H)}_{\dot{\rho}(H) / \rho(H)} - \underbrace{ \vphantom{\dv{}{t}}\big( \avgL - \L(\H) \big)}_{ \dot{\rho}^{\star}(H) / \rho(H)} \bigg)^{2}
    \right] =
    \frac{1}{2}
    \Big(\dot{\rho} - \dot{\rho}^{\star}\Big)^{i}
    \F_{ij}(\rho)
    \Big(\dot{\rho} - \dot{\rho}^{\star}\Big)^{j}.
  \end{aligned}
\end{equation}
\end{restatable}
As desired, \(\Err\) defines a distance in the tangent space of \(\HHH\),
imposed by the Fisher metric \(\F(\rho)\) between the replicator dynamics
\(\dot{\rho}^{\star}\) and the dynamics \(\dot{\rho}\) realized by \ngd with
respect to \(\theta\). By inspection, we see that \(\Err\) is minimized for
tabular settings (\(\rho \equiv \theta\)) if and only if the replicator equation
holds (\ie, \(\dot{\rho} = \rho ( \avgL - \L(\h) )\)). The minimization of
\(\Err\) by \ngd generalizes to any twice-differentiable parameterization of
\(\rho\) by \(\theta\).

\begin{restatable}[\textbf{Conjugate Natural Selection; Main Result}]{thm}{thmcns}\label{thm:cns}
Constrained to a given manifold of twice-differentiable parametric policies
\(\rho(h ; \theta)\), \ngd of \(\avgL\) with respect to \(\theta\)
(\cref{eq:ngd,eq:conj-ngd}) achieves the least-squares optimal fit in
\(\dot{\theta}\) to the continuous time replicator dynamics (\ie,
\cref{eq:replicator}), as measured by the natural deviation \(\Err\)
(\cref{defi:err}).
\end{restatable}

Our \cref{prf:cns}, provided in the supplementary material, proceeds by
establishing that \ngd of \(\avgL\) with respect to \(\theta\) induces a
stationary point of \(\Err\), such that \(\pt\Err/\pt\dot{\theta} = 0\), with a
Hessian that is positive semi-definite everywhere, implying a global minimum.

In addition to proving an optimal correspondence between \ngd with respect to
\(\theta\) and the continuous time replicator equation (\cref{thm:cns}), we may
characterize the space of functions \(u \colon \HH \to \RR\) that undergo the
same dynamics under either update rule as linear combinations of score
(\cref{thm:price}). Finally, we demonstrate an application of conjugate natural
selection (\cns) by experimentally evolving a distribution of continuous
hypotheses for a non-convex problem.
\begin{restatable}[Preserved Dynamics]{thm}{thmprice}\label{thm:price}
  Linear combinations of score satisfy the Price equation (\cref{eq:price}) when
  \(\theta\) is updated via \ngd of \(\avgL\). That is,
  \begin{equation} \label{eq:price-thm}
    \forall \alpha^{i} \in \RR, u = \alpha^{i} s_{i}(\theta ; \h),
    \quad \dv{}{t} \E_{\H \sim \rho}\Big[u(\H)\Big]
    = -\Cov_{\H\sim\rho}\Big[u(\H), \L(\H)\Big]
    + \E_{\H\sim\rho}\Big[\dot{u}(\H)\Big].
  \end{equation}
\end{restatable}

We include \cref{prf:price} in the supplementary material.

\subsection{Applications}\label{sec:cns-applications}

\begin{figure}[ht]
  \centering
  \begin{minipage}{0.20\textwidth}
    \centering
    \includegraphics[width=\textwidth]{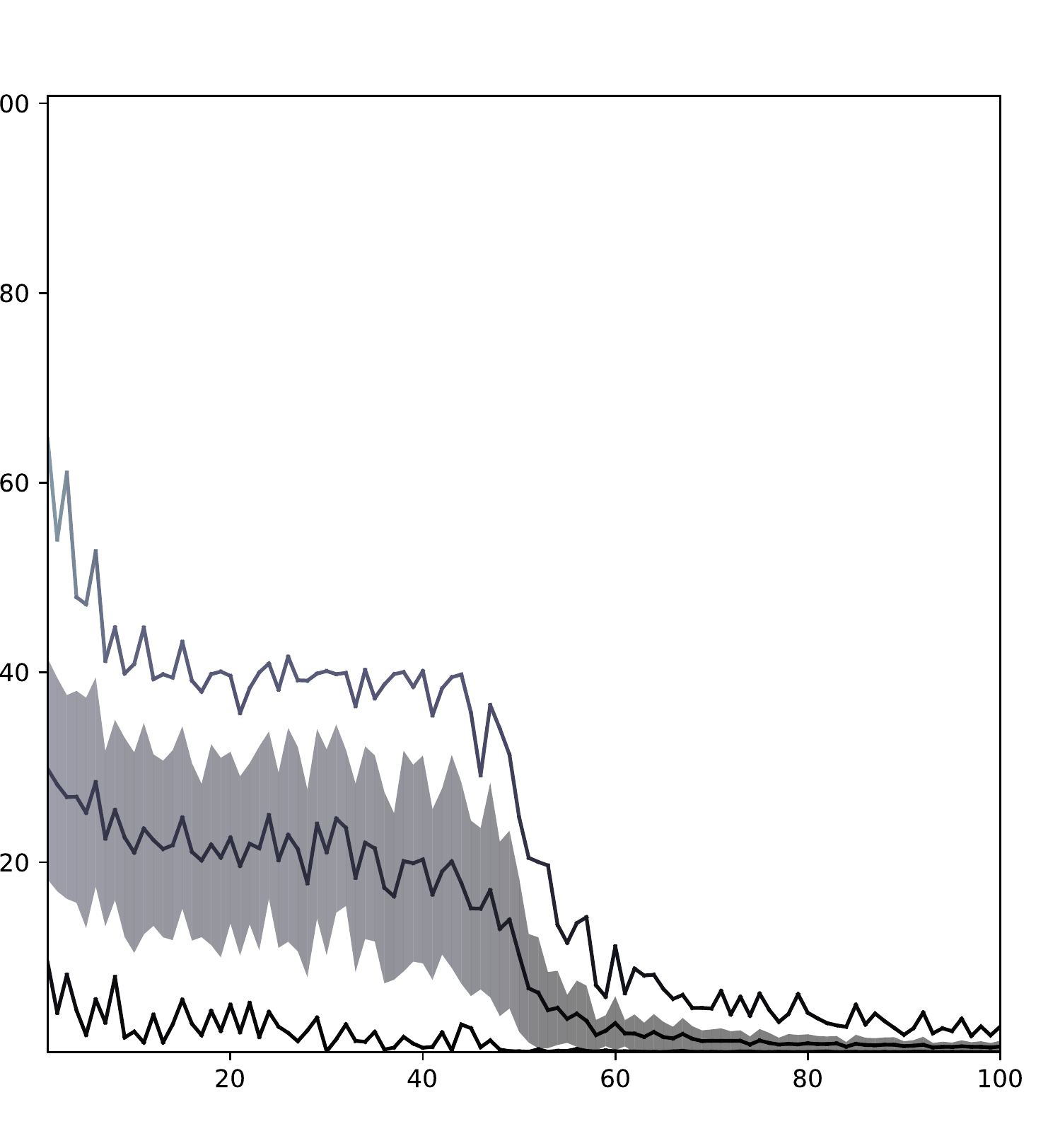}
    \tiny Empirical Losses vs Time
  \end{minipage}
  \begin{minipage}{0.15\textwidth}
    \centering
    \includegraphics[width=\textwidth]{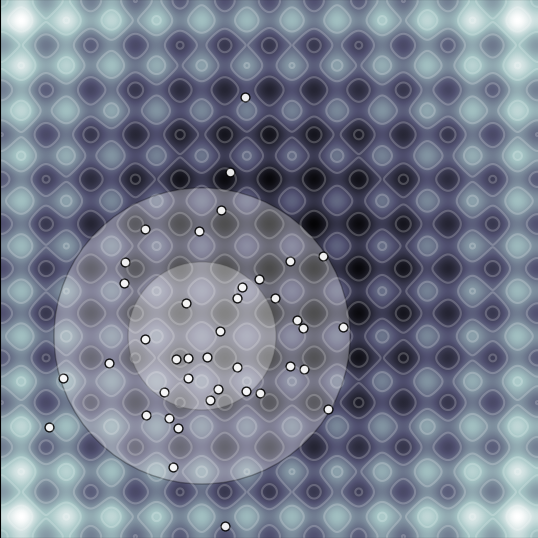}
    \tiny Step 0
  \end{minipage}
  \begin{minipage}{0.15\textwidth}
    \centering
    \includegraphics[width=\textwidth]{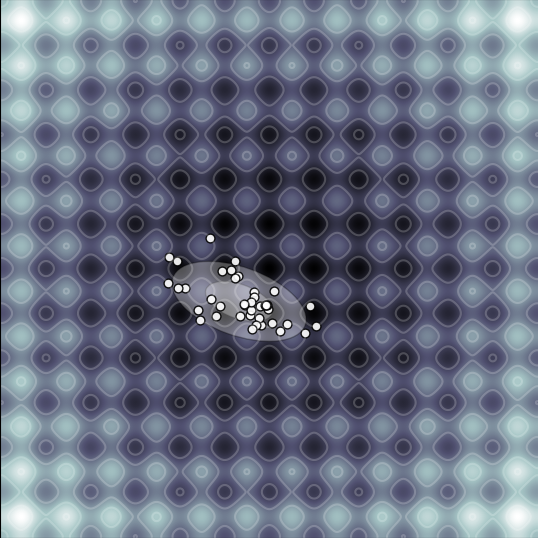}
    \tiny Step 20
  \end{minipage}
  \begin{minipage}{0.15\textwidth}
    \centering
    \includegraphics[width=\textwidth]{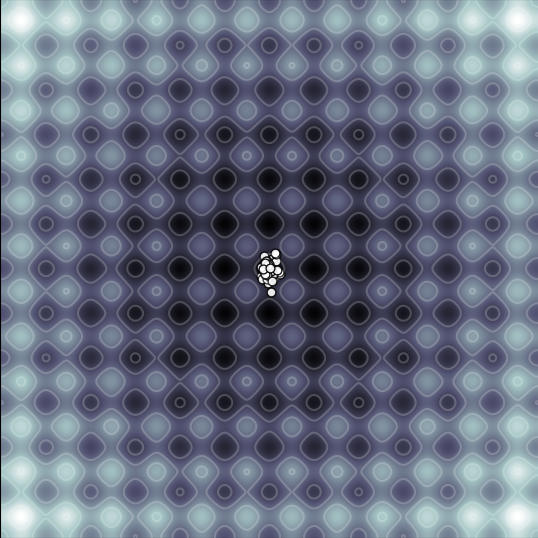}
    \tiny Step 50
  \end{minipage}
  \begin{minipage}{0.3\textwidth}
    \centering
    \includegraphics[width=\textwidth, trim=3cm 2.0cm 3cm 3cm, clip]{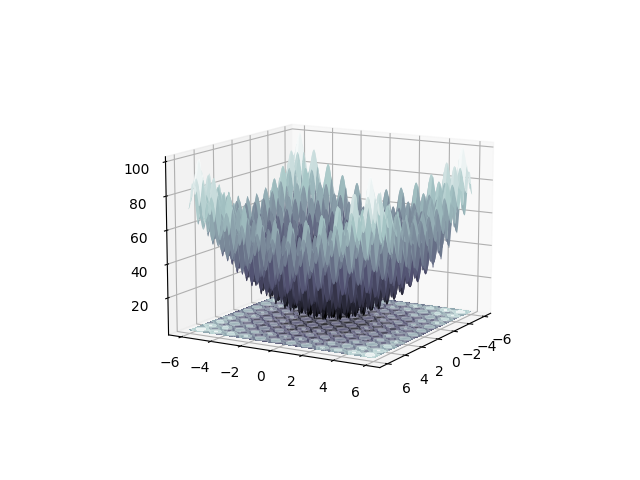}
    \tiny Rastrigin Loss Function \(\L(\h)\)
  \end{minipage}
  \caption{On the left, we plot the mean, standard deviation, and extremal
empirical losses for the learned distribution over 100 time steps. On the right,
the loss function is visualized as a surface over the domain
\([-6, 6] \times [-6, 6]\). In the middle, we represent time steps 0, 20, and 50
of the evolution: The Rastrigin function is visualized with shading and
highlighted level sets; the sampled hypothesis are represented by white dots;
and the 1- and 2-\(\sigma\) ellipses for the evolving Gaussian distribution
\(\rho\) are shaded white with partial transparency. The distribution is
initialized with mean at \(\mathtt{[-1.5, -1.5]}\) and identity covariance, and
we use a constant learning rate of \(\mathtt{1e^{-3}}\) for the Euler update. An
animation of the time-evolution of the distribution is available with the
included source code.}
\label{fig:rastrigin}
\end{figure}

We demonstrate an application of \cns by evolving a Gaussian distribution over
candidate solutions for a non-convex optimization problem: namely, unconstrained
minimization of the Rastrigin function,
\[
 \L(h_{x},h_{y}) = 20 + h_{x}^{2} + h_{y}^{2} - 10 \cos(2\pi h_{x}) - 10 \cos(2 \pi h_{y}),
\]
depicted in the rightmost pane of \cref{fig:rastrigin}.

At each time step, \(N{=}40\) hypotheses \(h\) are sampled from \(\rho_{t}\) and the
loss for each \(\h\) is calculated, yielding a Monte Carlo estimate of the loss
gradient
\( \pt_{i} \avgL \approx \frac{1}{N}\sum_{k=1}^{N} \L(\h_{k}) \pt_{i} \log \rho(\h_{k}) \).
We use an analytically-known form of the Fisher for a non-degenerate
parameterization of a 2-dimensional Gaussian distribution using 5 degrees of
freedom and an Euler discretization of the dynamics. We provide our code for
this simulation in the supplementary material.

\subsection{Limitations}
Recent characterizations of \ngd have suggested quadratic convergence rates
under certain conditions \citep{muller2022geometry, hu2022riemannian}, though
care must be taken to choose appropriate parametric manifolds \(\MM\) for
\(\L\). (Intuitively, it is possible to \qt{optimally} fit data to any model,
but the model must be appropriate to the domain for the optimal fit to be
useful).

Until recently, the bottleneck for applying \ngd in practice was the
\(\mathcal{O}(n^{\approx 2.37})\) cost of Fisher matrix inversion, prompting
alternative empirical approximations of natural gradient descent
\citep{martens2020new, hennes2019neural, peirson2022fishy}. A more scalable
approach, based on solving the corresponding least-squares problem directly, has
been recently proposed by \citet{nurbekyan2022efficient}.

Our simple demonstration in \cref{fig:rastrigin} indicates that conjugate
natural selection provides an alternative approach to standard approaches in
evolutionary computation: Rather than directly simulate a population, we may use
\ngd to update a parametric \emph{distribution} of candidate solutions and
ultimately solve a non-convex optimization problem, even when the hypothesis
space is continuous or high-dimensional. As \(\avgL\) and \(\F(\theta)\) are
often approximated by empirical averages, we also comment that this approach
readily extends in the presence of noise that is independent of \(\rho\) or
\(\theta\).  In particular, the hypothesis space \(\HH\) may correspond to the
space of \emph{functions} over a random input, as in \cref{sec:bayes}. Finally,
we assert that using \ngd for evolutionary computation may be suitable for
\emph{constrained} optimization in practice, because simple sample rejection can
guarantee that domain constraints for \(\h\) are satisfied (although sample
rejection will distort the corresponding Fisher information matrix).


\section{Continuous Bayesian Inference} \label{sec:bayes}

For this section, let us interpret \(h\) as a predictive model yielding a
probability (density) for a stochastic process \(\X_{t}\) observed in continuous
time \(t\) and distributed by Nature as \(\n\).  Examples of such processes
\(X_{t}\) include physical quantities like instantaneous field amplitude; the
idealized market price of an asset; Brownian motion; or any continuous time
quantity perturbed by \emph{noise}.  In this context, the replicator equation
(\cref{eq:replicator}) corresponds to \emph{continuous Bayesian inference} if we
identify loss with the negative log-likelihood of hypothesis \(\h\) given
\(\x_{t}\).  That is, let
\begin{equation}\label{eq:surprisal}
  \h(x_{t} ; t) = \Pr_{\h}(\X_{t}{=}\x_{t} \smid t)
  \quad ; \quad
  \n(x_{t} ; t) = \Pr_{\n}(\X_{t}{=}\x_{t} \smid t)
  \quad ; \quad
  \L(\h, t) = -\log \h(\x_{t} ; t).
\end{equation}
The loss \(\L\) expressed in \cref{eq:surprisal} corresponds to
\emph{surprisal}, or the amount of information about \(X_{t}\) revealed under
hypothesis \(\h\) by the observation \(X_{t}{=}\x_{t}\). A good hypothesis
minimizes average surprisal by correctly predicting the process \(X_{t}\). For
this loss, the replicator equation (\cref{eq:replicator}) describes stochastic
dynamics for \(\rho\) that depend on the instantaneous value of \(\x_{t}\):
\begin{equation}
  \label{eq:bayes}
  \dot{\rho}_{t}(\h, \x_{t}) = \rho_{t}(\h) \Big[ \avgL_{\rho_{t}}(x_{t}) + \log \h(\x_{t} ; t) \Big],
  \quad \text{where} \quad
\avgL_{\rho_{t}}(x_{t}) = -\sum_{\h} \rho_{t}(\h) \log \h(\x_{t} ; t).
\end{equation}
We will use the gradient of the expected value of \(\avgL_{\rho_{t}}(\cdot)\) over
\(\X_{t} \sim \n\) and perform \ngd to evolve \(\rho\). While
\(\avgL_{\rho_{t}}(\cdot)\) might be formally defined as a cross-entropy term,
its \emph{gradient} with respect to \(\rho_{t}\) is the same as the gradient of
the expected Kullback-Leibler divergence (relative-entropy) \(\DD\) from \(h\) to
\(\n\), defined
\begin{equation}\label{eq:kl}
  \DD_{t}(\n \parallel \h) \define -\sum_{x} \n(\x ; t) \log \frac{\h(\x ; t)}{\n(\x ; t)}.
\end{equation}
\begin{restatable}[]{lem}{lemgradients}\label{lem:gradients}
  The gradients of
  \(\E_{\X_{t}\sim\n}[\avgL_{\rho_{t}}(\X_{t})]\)
  and
 \(\E_{\H\sim\rho_{t}}[\DD_{t}(\n \parallel \H)]\)
 with respect to \(\rho_{t}\) are equal.
\end{restatable}
A proof of \cref{lem:gradients} is provided in the supplementary material.

\begin{restatable}[Continuous Inference]{thm}{thmbayes}\label{thm:bayes}
  \cref{eq:bayes} may be used to derive Bayes's rule.
\end{restatable}
We provide \cref{prf:bayes} in the supplementary material. Unfortunately, when
\(\HH\) is large, it is \textbf{difficult to compute the \emph{prior}} for an
observable, as this calculation requires integrating over \(\HH\) (\ie,
\(\Pr_{\rho_{t}}(X_{t}^{\Delta t}) = \sum_{h} \rho_{t}(\h) \h(X_{t}^{\Delta t} , t)\)
in \cref{eq:bayes-discrete}). This difficulty is used to justify approximate
Bayesian inference based on variational bounds, such as the Evidence Lower Bound
(ELBO). We may \textbf{avoid the corresponding difficulty via \ngd}, however, by
using Monte Carlo sampling to estimate the necessary gradient,
\( \pt_{i} \E_{\H \sim \rho}[\DD_{t}(\n \parallel \H)] = -\E_{X_{t},H}\left[\pt_{i} \log \rho(\H ; \theta) \log \H(\X_{t} , t) \right]\).

\begin{restatable}[\ngd Yields Optimal Continuous Inference]{thm}{thmoci}\label{thm:oci}
 For any probability distribution \(\rho(\h ; \theta)\) that is
twice-differentiable with respect to parameters \(\theta\), \ngd of the expected
divergence \(\E_{\H \sim \rho_{t}}[\DD_{t}(\n \parallel \H)]\) (of the
\(\rho\)-weighted predictions of model \(\h(\X_{t} ; t)\) for \(X_{t} \sim \n\)) with respect to \(\theta\)
optimally approximates Bayesian inference for \(\rho\) in continuous time, by
minimizing \(\Err\) (\cref{defi:err}).
\end{restatable}

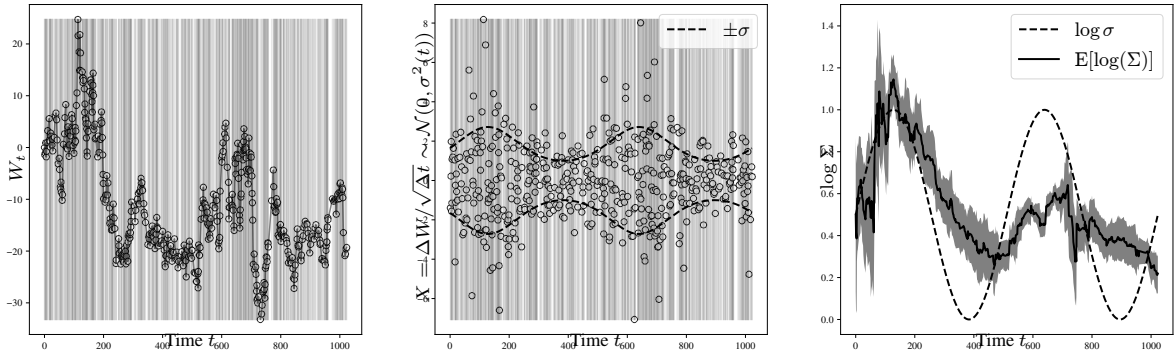
\begin{figure}[ht]
  \centering
  \begin{minipage}{0.30\textwidth}
    \begin{tikzpicture}
      \node (img) {\includegraphics[width=\textwidth, clip, trim=1cm 1cm 1cm 1cm]{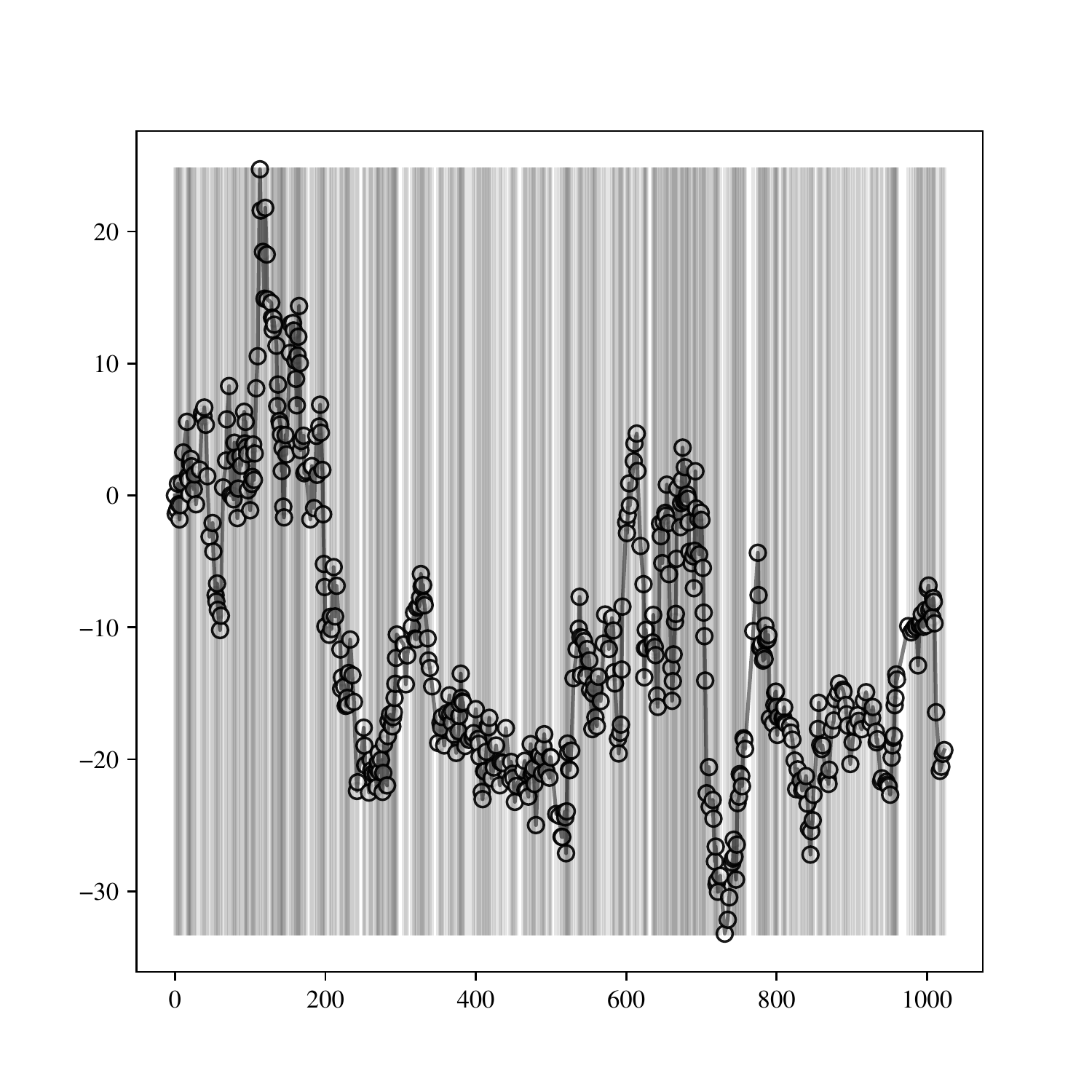}};
      \node[yshift=-2.3cm] {\scriptsize Time \(t\)};
      \node[rotate=90, yshift=2.3cm] {\scriptsize \(W_{t}\)};
    \end{tikzpicture}
  \end{minipage}
  \quad
  \begin{minipage}{0.30\textwidth}
    \begin{tikzpicture}
      \node (img) {\includegraphics[width=\textwidth, clip, trim=1cm 1cm 1cm 1cm]{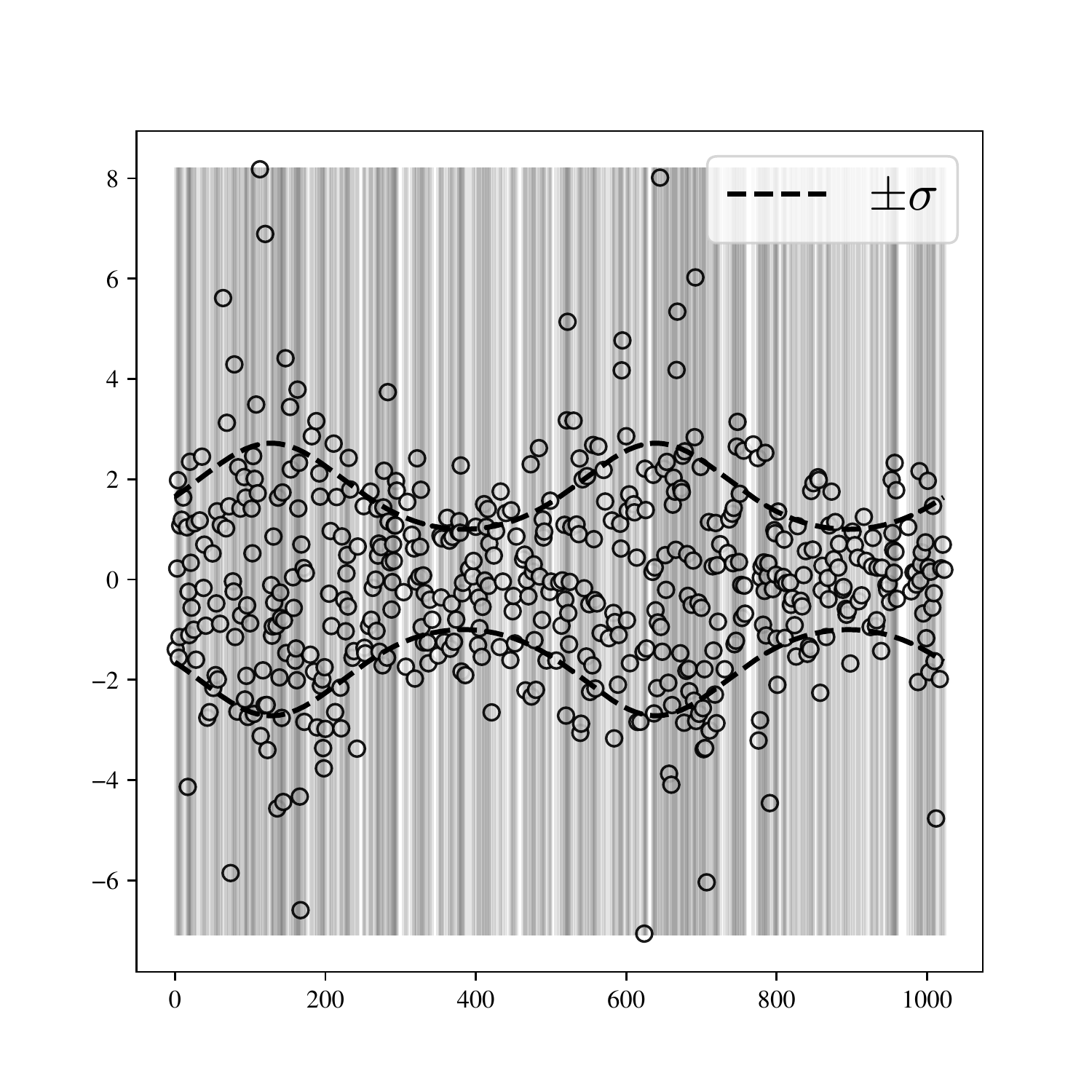}};
      \node[yshift=-2.3cm] {\scriptsize Time \(t\)};
      \node[rotate=90, yshift=2.3cm] {\scriptsize \(\X = \Delta W/\sqrt{\Delta t} \sim \mathcal{N}(0, \sigma^{2}(t))\)};
    \end{tikzpicture}
  \end{minipage}
  \quad
  \begin{minipage}{0.30\textwidth}
    \begin{tikzpicture}
      \node (img) {\includegraphics[width=\textwidth, clip, trim=1cm 1cm 1cm 1cm]{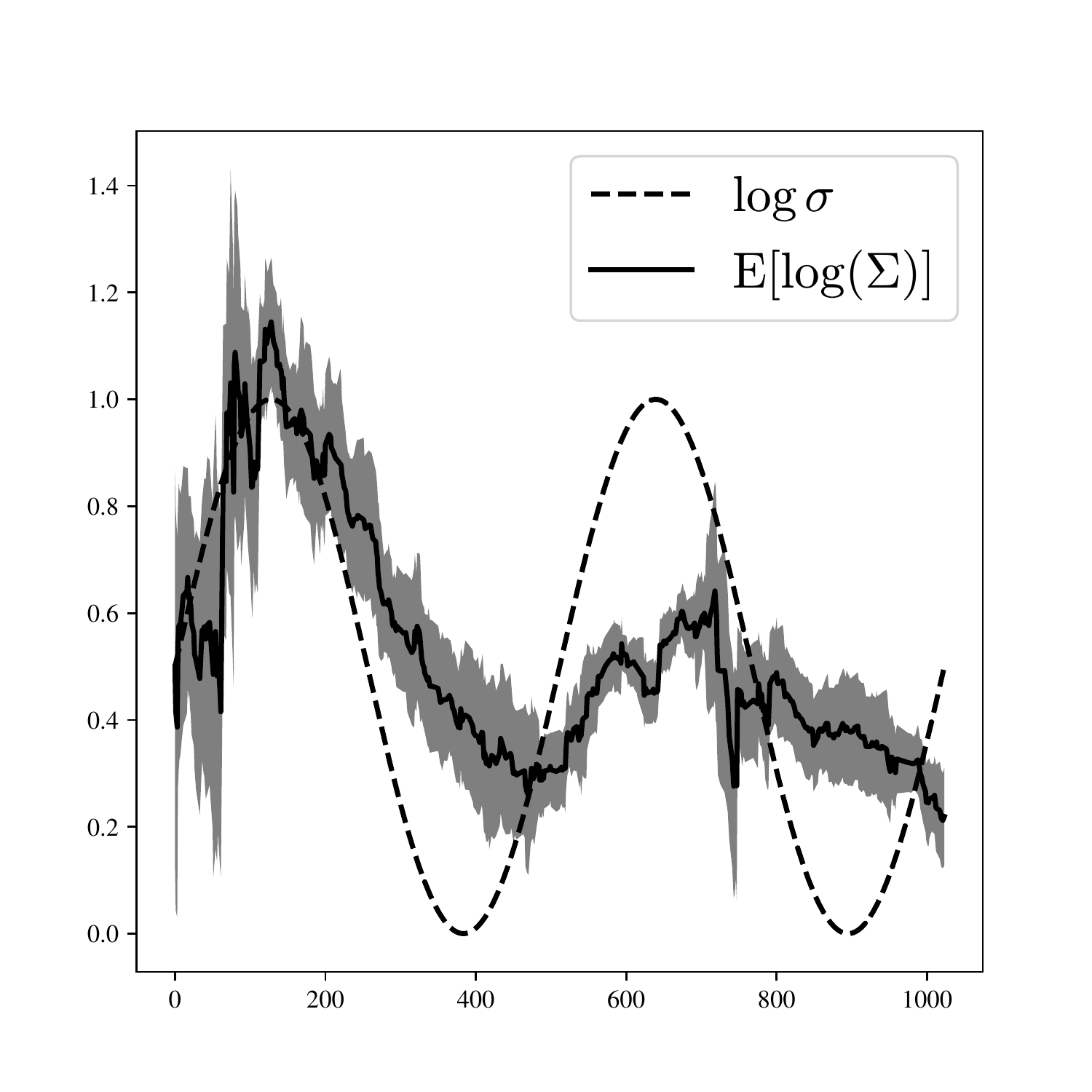}};
      \node[yshift=-2.3cm] {\scriptsize Time \(t\)};
      \node[rotate=90, yshift=2.3cm] {\scriptsize \(\log \Sigma\)};
    \end{tikzpicture}
  \end{minipage}
  \caption{ Given randomly-spaced samples of a Wiener process \(W_{t}\) with
time-dependent variance (left), where vertical lines indicate the times of
observed samples (left, center), we define the observable
\(\X_{t} = \Delta W/\sqrt{\Delta t} \sim \n = \mathcal{N}(0, \sigma^{2}(t))\)
between measurements of \(W_{t}\) (center). We update a \(\theta\)-parameterized
Gaussian distribution \(\rho\) over \(\H = \log \Sigma\) via \ngd, with respect
to \(\theta\), of the expected Kullback-Leibler divergence between \(n\) and
\(\H \sim \rho\), interpreting \(\Sigma \) as an estimate of the time-evolving
parameter \(\sigma(t)\). We visualize the mean and standard-deviation of
\(\log \Sigma \sim \rho\) with time, compared to the actual value of
\(\log \sigma\) (right).}
\label{fig:wiener}
\end{figure}

\subsection{Parameter Estimation for a Stochastic Process}\label{sec:oci-applications}

Using \ngd, we demonstrate learning a time-varying \emph{distribution} \(\rho = \mathcal{N}(\theta; t)\)
over \(\H = \log(\Sigma)\), where \(\Sigma\) is an
estimator for the time-varying parameter \(\sigma(t)\) of an observable Wiener
process
\(
  \dd{W}_{t} \sim \mathcal{N}(0, \sigma^{2}(t)\dd{t})
\)
(\cref{fig:wiener}).
Our example uses a 40-sample Monte Carlo gradient estimate and an Euler
discretization of the dynamics
\(\theta_{t+1} = \theta_{t} + \eta \dot{\theta}_{t}\) with constant learning
rate \(\eta = \mathtt{1e^{-2}}\).

\section{Conclusion}
We have shown that \ngd optimally approximates evolutionary and Bayesian
dynamics for any twice-differentiable parameterization of a distribution over
hypotheses. We believe it is remarkable that the essential dynamics of evolution
by natural selection share such close relationships with the fundamentals of
information theory, and that the unifying theoretical machinery is widely
applicable to machine learning and optimization in practice.

In the case of the correspondence between \ngd and evolutionary dynamics, we
have termed our finding \qt{conjugate natural selection} and demonstrated its
application to a non-convex optimization problem over a continuous hypothesis
space. We assert that this approach provides an alternative to existing methods
of evolutionary computation by dispensing with the need to simulate populations


\paragraph{Acknowledgments} This work is partially supported by the National
Science Foundation (NSF) under grants IIS-2143895 and IIS-2040800 (FAI program
in collaboration with Amazon), and CCF-2023495. We thank Shadi Haddad, Yin Lin,
Andrew Warren, Warren Mardoum, Ehsan Amid, and Abhishek Halder for feedback on
this or prior versions of our manuscript or for consultation on technical
details.

\bibliographystyle{plainnat}
\bibliography{references.bib}

\newpage
\input{appendix.tex}
\end{document}

%% file: macros.tex
\usepackage[english]{babel} 

\usepackage[utf8]{inputenc} 
\usepackage[T1]{fontenc}    

\usepackage{amsfonts}       
\usepackage{nicefrac}       
\usepackage{microtype}      
\usepackage{dsfont}         
\usepackage{mathrsfs}       
\usepackage{bm}             
\usepackage{xcolor}         
\usepackage{xspace}         

\usepackage[]{amsthm, thmtools, thm-restate}
\usepackage[]{amssymb}      
\usepackage[]{mathtools}    

\usepackage{physics}

\usepackage{cancel}         
\usepackage{centernot}      

\usepackage{booktabs}       
\usepackage{graphicx}       
\usepackage[]{tikz}         
\usepackage[]{pgfplots}     
\usepackage{adjustbox}

\usepackage{float}          
\usepackage{caption}
\usepackage{wrapfig}        

\usepackage{url}                               
\usepackage{hyperref}                          
\hypersetup{
  colorlinks, citecolor=teal, linkcolor=blue
}

\usepackage{nameref}                           
\usepackage[capitalize, nameinlink]{cleveref}  
\usepackage{crossreftools}
\crefformat{footnote}{#2\footnotemark[#1]#3}          

\crefformat{problem}{#2 Prob.~(#1)#3}

\newcommand\scalemath[2]{\scalebox{#1}{\mbox{\ensuremath{\displaystyle #2}}}}

\crefname{Problem}{Problem}{Problem}

\newcommand{\define}{\coloneqq}

\newcommand{\avg}[1]{\overline{#1}}
\newcommand{\wtilde}[1]{\widetilde{#1}}
\DeclareMathOperator*{\E}{E} 
\DeclareMathOperator*{\HH}{H}  
\DeclareMathOperator*{\Cov}{Cov}  

\DeclareMathOperator*{\argmin}{argmin}

\newcommand{\RR}{\mathbf{R}} 

\newcommand{\eg}{{e.g.}\xspace}  
\newcommand{\ie}{{i.e.}\xspace}  
\newcommand{\etc}{{etc.}\xspace} 

\newcommand{\qt}[1]{``#1''}

\AtBeginDocument{}

\declaretheorem[name=Lemma, style=plain,
  refname={Lem.,Lems.},
  Refname={Lemma,Lemmas}]{lem}

\declaretheorem[name=Definition, style=definition,
  refname={Def.,Defs.},
  Refname={Definition,Definitions}]{defi} 

\declaretheorem[name=Remark, sharenumber=lem, style=plain,
  refname={Rem.,Rems.},
  Refname={Remark,Remarks}]{rem}

\declaretheoremstyle[
    spaceabove=-6pt,
    spacebelow=6pt,
    headfont=\normalfont\itshape,
    bodyfont = \normalfont,
    postheadspace=1em,
    qed=$\square$,
    headpunct={.}]{myproofstyle} 
\declaretheorem[name={Proof}, style=myproofstyle]{ProofINNER}

\newenvironment{Proof}[1]
 {\renewcommand{\theProofINNER}{#1}\ProofINNER}
 {\endProofINNER}
\crefformat{ProofINNER}{#2Proof~#1#3}

%% file: appendix.tex
\appendix

\section{Deferred Proofs}

We organize our deferred proof thematically. In \cref{sec:usefullemmas}, we
provide lemmas that allow for more succinct proofs throughout the rest of this
section.  In \cref{sec:eqproj}, we derive \cref{eq:proj}, appearing in the main
text.  In \cref{sec:prf-cns}, we prove our primary result, deemed ``conjugate
natural selection'' (\cref{thm:cns}). In \cref{sec:prf-price}, we derive the
Price equation and our result regarding dynamics preserved between replicator
dynamics and \ngd on \(\MM\) (\cref{thm:price}).  Finally, in
\cref{sec:prf-bayes}, we derive the discrete replicator equation and Bayes's
rule, and we prove the optimality of \ngd for continuous Bayesian inference
(\cref{thm:bayes}).

\subsection{Four Useful Lemmas (5-8)} \label{sec:usefullemmas}

\cref{lem:functional,lem:conj-flow,lem:conj-loss} provide equivalent expressions
for quantities that frequently appear in our problem domain:
\cref{lem:functional} identifies a simple term-rewriting rule, while
\cref{lem:conj-flow} and \cref{lem:conj-loss} address either side of the
conjugate gradient flow equation (\cref{defi:conj-ngd}), restated below.
\cref{lem:dots} proves a simple identity relying on repeated application of the
chain rule.
\begin{lem}
  \label{lem:functional}
  \emph{(Functionals)}.  For any function \(u \colon \HH \to \RR\), for all
parameter components \(i\), the following expressions are equivalent:
\[
  \sum_{\h} \pt_{i} \rho(\h) u(\h)
  = \E_{\H\sim\rho}\left[
   u(\H) \pt_{i} \log \rho(\H)
  \right]
  = \Cov_{\H\sim\rho} \left[
    \pt_{i} \log \rho(\H), u(\H)
  \right].
\]
\end{lem}
\begin{proof}
  The first equality is an instance of the ``log-derivative trick'', while the
second follows from \cref{lem:score}:
  \[
    \forall u, \quad \sum_{\h} \pt_{i}\rho(\h) u(\h)
    =
    \underbrace{\sum_{\h} \rho(\h) \pt_{i} \log \rho(\h) u(\h)}_{
\E_{\H\sim\rho}\left[
   u(\H) \pt_{i} \log \rho(\H)
  \right]
    }
    = \Cov_{\H\sim\rho} \left[\pt_{i} \log \rho(\H), u(\H) \right].
  \]
\end{proof}

Briefly recall the definitions for the Fisher (\cref{defi:fisher}) and conjugate
gradient flow under \ngd (\cref{defi:conj-ngd}), which we reference in our proof
of \cref{lem:conj-flow}:

\defifisher*
\deficonjngd*

\begin{lem}
  \label{lem:conj-flow}
  \emph{(Conjugate Flow)}
  The dynamics of \(\rho \in \MM\) under \ngd of \(\avgL\) with respect
to \(\theta\) (\cref{defi:conj-ngd} are
  \begin{equation*}
  \F_{ij}(\theta) \dot{\theta}^{j}
  = \Cov_{\H \sim \rho} \Big[ \pt_{i} \log \rho(\H),~ \dv{}{t} \log \rho(\H) \Big]
  = \sum_{\h}  \Big[\pt_{i} \log \rho(h)\Big] \dot{\rho}(\h).
\end{equation*}
where we elide the explicit dependence of \(\rho\) on \(\theta\) for compactness.
\end{lem}
\begin{proof}
The first equality follows from substituting the definition of Fisher
information (\cref{defi:fisher}) into \cref{defi:conj-ngd} and summing over
\(j\), by the linearity of covariance.
\begin{align*}
  \F_{ij}(\theta) \dot{\theta}^{j}
  &= \Cov_{\H \sim \rho} \Big[ \pt_{i} \log \rho(\H),~ \pt_{j} \log \rho(\H) \Big] \dot{\theta}^{j} \\
  &= \Cov_{\H \sim \rho} \Big[ \pt_{i} \log \rho(\H),~ \dv{}{t} \log \rho(\H) \Big].
\end{align*}
The second follows from rewriting
covariance as an explicit sum over \(\h\) while invoking \cref{lem:score} and
pushing the time derivative through the logarithm, that is,
\begin{align*}
  \Cov_{\H \sim \rho} \Big[ \pt_{i} \log \rho(\H),~ \dv{}{t} \log \rho(\H) \Big]
  &= \Cov_{\H \sim \rho} \Big[ \pt_{i} \log \rho(\H),~ \frac{1}{\rho(\H)} \dv{}{t} \rho(\H) \Big] \\
  &= \sum_{\h} \rho(\h) \Big[ \pt_{i} \log \rho(\h) \Big] \frac{1}{\rho(\h)} \dv{}{t} \rho(\h) \\
  &= \sum_{\h}  \Big[ \pt_{i} \log \rho(\h) \Big] \dot{\rho}(\h).
\end{align*}
\end{proof}
\begin{lem}
  \label{lem:conj-loss}
  \emph{(Loss Gradient)}
  The gradient of expected loss \(\avgL\) with respect to \(\theta\) is
  \begin{align*}
  \pt_{i} \avgL
  = \Cov_{\H \sim \rho} \Big[ \pt_{i} \log \rho(\H), ~ \L(\H) \Big]
  = - \sum_{\h} \Big[ \pt_{i} \log \rho(\h) \Big] \rho(\h) \Big(\avgL -  \L(\h) \Big).
  \end{align*}
\end{lem}
\begin{proof}
  As the function \(\L \colon \HH \to \RR\) has no explicit dependence on
\(\theta\), we may expand
  \[
    \pt_{i} \avgL = \sum_{h} \L(\h) \pt_{i} \rho(\h).
  \]
  By \cref{lem:functional}, when we choose \(u = \L\), we may equate this sum
with the desired covariance between \(\pt_{i} \log \rho\) and \(\avgL\):
  \[
    \sum_{h} \L(\h) \pt_{i} \rho(\h)
    = \Cov_{\H\sim\rho}[ \pt_{i} \log \rho(\H), \L(\H)].
  \]
  Finally, we re-express this covariance as an explicit sum:
  \[
\Cov_{\H \sim \rho} \Big[ \pt_{i} \log \rho(\H), ~ \L(\H) \Big]
  = - \sum_{\h} \Big[ \pt_{i} \log \rho(\h) \Big] \rho(\h) \Big(\avgL -  \L(\h) \Big).
  \]
\end{proof}

\begin{lem}
  \label{lem:dots}
  \emph{(Cancel Dots with Chain Rule)}. \(\forall u\) independent of
\((\dot{\theta}, \dot{\rho})\),
\[
  \pdv{}{\dot{\theta}^{i}}\left( \dv{}{t} u(\rho) \right) = \pt_{i} u(\rho).
\]
\end{lem}
\begin{proof}
  By the chain rule,
  \begin{equation*}
    \scalemath{0.98}{
\pdv{}{\dot{\theta}^{i}}\left( \dv{}{t} u(\rho) \right)
= \pdv{}{\dot{\theta}^{i}} \left( u'(\rho) \dot{\rho} \right)
= u'(\rho) \pdv{\dot{\rho}}{\dot{\theta}^{i}}
= u'(\rho) {\pdv{}{\dot{\theta}^{i}}} \left( \dot{\theta}^{i} \pt_{i}\rho + \pt_{t} \rho \right)
= u'(\rho) \pt_{i} \rho
= \pt_{i} u(\rho)}.
\end{equation*}
\end{proof}

\subsection{Proof: Projection of Replicator Dynamics (\cref*{eq:proj})} \label{sec:eqproj}

In the main text, we claim that the conjugate gradient flow (\cref{defi:conj-ngd})
may be re-written as

\begin{equation*}
  \sum_{\h}  \Big[\underbrace{\pt_{i} \log \rho(h)}_{s_{i}}\Big] \dot{\rho}(\h)
  = \sum_{\h} \Big[ \underbrace{\pt_{i} \log \rho(\h)}_{s_{i}}\Big] \rho(\h) \Big(\avgL -  \L(\h) \Big). \tag{\ref*{eq:conj-ngd}}
\end{equation*}

\begin{proof}
  By direction application of \cref{lem:conj-flow} and \cref{lem:conj-loss},
  we compare \cref{eq:proj} to \cref{eq:conj-ngd}:
  \begin{align*}
    \F_{ij}(\theta) \dot{\theta}^{j}
    &= - \pt_{i} \avgL. \\
    \sum_{\h}  \Big[\pt_{i} \log \rho(h)\Big] \dot{\rho}(\h)
    &= \F_{ij}(\theta) \dot{\theta}^{j}. \\
    - \pt_{i} \avgL
    &= \sum_{\h} \Big[ \pt_{i} \log \rho(\h) \Big] \rho(\h) \Big(\avgL -  \L(\h) \Big). \\
    \therefore \quad
    \sum_{\h}  \Big[\pt_{i} \log \rho(h)\Big] \dot{\rho}(\h)
   &= \sum_{\h} \Big[\pt_{i} \log \rho(\h)\Big] \rho(\h) \Big(\avgL -  \L(\h) \Big).
  \end{align*}
\end{proof}

\subsection{Proof: Conjugate Natural Selection} \label{sec:prf-cns}

With \cref{lem:first-derivative}, we establish that \ngd of \(\avgL\) with
respect to \(\theta\) (\cref{defi:conj-ngd}) induces a local extremum of
\(\Err\) (\cref{defi:err}) with respect to \(\dot{\theta}\). With
\cref{lem:second-derivative}, we establish that this local extremum is a global
minimum, by the fact that the Hessian of \(\Err\) with respect to
\(\dot{\theta}\) is everywhere positive semi-definite; In fact, this Hessian is
the Fisher. We conclude that \ngd of \(\avgL\) with respect to \(\theta\) is an
optimal approximation of the replicator dynamics, a finding we term ``conjugate
natural selection'' (\cref{thm:cns}).

We first restate \cref{defi:err}:

\defierr*

We additionally recall that \(\F(\rho)\) is given by
\begin{equation}
  \F_{ij}(\rho) = \E_{\H\sim\rho}\left[\frac{\delta_{ij}}{\rho(H)^{2}}\right].
  \tag{\ref*{eq:shahshahani}}
\end{equation}

\begin{lem}
  \label{lem:first-derivative}
  (Gradient of Natural Deviation)
  \[
     \pdv{\dot{\theta}^{i}}\Err(\dot{\theta}) =
    \F_{ij}(\theta) \dot{\theta}^{j} + \pt_{i} \avgL.
  \]
\end{lem}
\begin{proof}
We may take the gradient of \cref{eq:err} with respect to \(\dot{\theta}\).
We use \cref{lem:dots} (for \(u = \log\)) to first write
\begin{equation*}
  \pdv{}{\dot{\theta}^{i}}\left( \dv{}{t} \log \rho(\h)  \right) = \pt_{i} \log \rho(\h),
\end{equation*}
thus,
\begin{equation} \label{eq:first-derivative}
  \pdv{\dot{\theta}^{i}}\Err(\dot{\theta})
     = \E_{\H\sim\rho}\left[
      \left(
      \dv{}{t} \log \rho(\H)
      - \left( \avgL - \L(\H) \right)
      \right) \pt_{i} \log \rho(\H)
    \right].
\end{equation}
  Recognizing that \(\dv{}{t} \log \rho = \frac{1}{\rho} \dot{\rho} \), the
expectation value on the right side of \cref{eq:first-derivative}
separates into two explicit sums; \ie,
  \[
  \pdv{\dot{\theta}^{i}}\Err(\dot{\theta}) =
    \sum_{\h}  \left[\pt_{i} \log \rho(h)\right] \dot{\rho}(\h)
    - \sum_{\h} \left[ \pt_{i} \log \rho(\h) \right] \rho(\h) \left(\avgL -  \L(\h) \right).
  \]
  These sums correspond to the gradient flow on \(\MM\)
(\cref{lem:conj-flow}) and the loss gradient
(\cref{lem:conj-loss}), respectively.
\end{proof}

\begin{lem}
  \label{lem:second-derivative}
  \emph{(Hessian of Natural Deviation)} \(\Err\) is convex in \(\dot{\theta}\),
that is,
   \[
     \pdv[2]{}{\dot{\theta}^{i}}{\dot{\theta}^{j}} \Err(\dot{\theta})
          = \F_{ij}(\theta) \succeq 0,
   \]
  where \(\F_{ij}(\theta) \succeq 0\) denotes that \(\F_{ij}(\theta)\) is positive
semi-definite (has only non-negative eigenvalues).
\end{lem}
\begin{proof}
We differentiate \cref{eq:first-derivative} with respect to \(\dot{\theta}\),
again using \cref{lem:dots} (for \(u = \log\)). Thus, the second derivative of
\(\Err\) is
\begin{align*}
     \pdv[2]{}{\dot{\theta}^{i}}{\dot{\theta}^{j}} \Err(\dot{\theta})
          &= \E_{\H\sim\rho} \Big[
          \left( \pt_{i} \log \rho(\H) \right)
          \left( \pt_{j} \log \rho(\H) \right)
          \Big] \\
          &= \Cov_{H \sim \rho}\Big[ \pt_{i} \log \rho(H ; \theta),~ \pt_{j} \log \rho(H ; \theta) \Big] \\
          &= \F_{ij}(\theta).
\end{align*}
Where the last equality relies on \cref{defi:fisher}.  As a covariance matrix,
\(\F\) is positive semi-definite (\ie, \(\F \succeq 0\)).
\end{proof}
That the Fisher is the Hessian of \(\Err\) with respect to \(\dot{\theta}\) is
unsurprising, since \(\Err\) is ultimately a distance measured by the Fisher
metric in the tangent space of \(\rho\). This underlying reason is shared with
characterizations of the Fisher as the Hessian of the loss surface
\citep{martens2020new}.

\thmcns*

\begin{Proof}{of \cref*{thm:cns}}\label{prf:cns}
  \cref{lem:first-derivative} implies that \ngd of \(\avgL\)
(\cref{defi:conj-ngd}) with respect to \(\theta\) achieves a local extremum of
\(\Err\) (\ie, $  \pdv{\dot{\theta}^{i}}\Err(\dot{\theta}) = 0$), while
convexity (\cref{lem:second-derivative}) guarantees that any local extremum of
\(\Err\) is a global minimum.
\end{Proof}

\subsection{Proof: The Price Equation and Preserved Dynamics}\label{sec:prf-price}

\cref{thm:price} characterizes the subspace of properties \(u(\h)\) that obey
the Price equation under \ngd: linear combinations of the score. This result
provides a direct route for determining how an approximation of the replicator
dynamics by \ngd in \(\theta\) for some chosen parameterization affects
quantities of interest: if a property is naturally expressed as a linear
combination of score, there is no resultant distortion of the dynamics of the
property in question when using a lower-dimensional representation \(\theta\)
with \ngd when compared to replicator dynamics in \(\HHH\).

We first restate the replicator equation (\cref{eq:replicator}) for local reference:
\begin{equation*}
  \dot{\rho}(\h) = \rho(\h) \Big[ \avgL_{\rho} - \L(\h) \Big],
  \quad\text{where}\quad
  \avgL_{\rho} \define \sum_{\h} \rho(\h) \L(\h), \quad \sum_{\h} \rho(\h) = 1. \tag{\ref*{eq:replicator}}
\end{equation*}

Next, we restate and prove \cref{lem:price}, defining the price equation, before
proving our characterization of the space of properties preserved by \ngd on
\(\MM\) when compared to the replicator dynamics:

\lemprice*
\vspace{1em}

\begin{Proof}{of \cref*{lem:price}} \label{prf:lemprice}
By the chain rule,
\[
  \forall u, \quad
  \dv{}{t} \sum_{\h} \rho(\h) u(\h)
  = \sum_{\h} \dot{\rho}(\h) u(\h) + \sum_{\h} \rho_{\h} \dot{u}(\h).
\]
Expanding \(\dot{\rho}(\h)\) in terms of the replicator equation
(\cref{eq:replicator}) and recognizing the terms of the equation as expectation
values, we have that
\[
  \forall u, \quad
  \dv{}{t}
  \underbrace{\sum_{\h} \rho(\h) u(\h)}_{
    \avgu_{\rho}
  }=
  \underbrace{\sum_{\h} \rho(\h)\Big[\avgL_{\rho} - \L(\h)\Big] u(\h)}_{
    -\Cov_{\H\sim\rho}\left[ u(\H), \L(\H) \right]
  }
  +
  \underbrace{\sum_{\h} \rho(\h) \dot{u}(\h)}_{
    \E_{\H\sim\rho}\left[ \dot{u}(\H) \right]
  }.
\]
\end{Proof}

\thmprice*
\vspace{1em}
\begin{Proof}{of \cref*{thm:price}}\label{prf:price}
  Differentiate \(\E[u] = \sum_{h}\rho(h)u(h)\) by the chain rule,
  where \(\dv{\rho}{t} = \rho \dv{}{t} \log \rho \) implies that
  \begin{equation} \label{eq:price-chain-log}
    \forall u, \quad \dv{}{t} \E_{\H\sim\rho} \Big[
    u(\H)
    \Big]
    = \E_{\H\sim\rho} \Big[
    u(H) \dv{}{t} \log \rho(\H)
    \Big]
    + \E_{\H\sim\rho} \Big[
    \dot{u}(\H)
    \Big].
  \end{equation}
  Independently, note that \cref{lem:conj-flow} and \cref{lem:conj-loss} allow us to
  rewrite (\cref{defi:conj-ngd}) as
 \begin{equation}\label{eq:conj-cov}
  \Cov_{\H \sim \rho} \Big[ \pt_{i} \log \rho(\H),~ \dv{}{t} \log \rho(\H) \Big]
  = -\Cov_{\H \sim \rho} \Big[ \pt_{i} \log \rho(\H), ~ \L(\H) \Big].
 \end{equation}
  When
\(u(\h) = \alpha^{i} s_{i}(\theta ; \h) = \alpha^{i} \pt_{i} \log \rho(\h)\) for
some vector \(\alpha(t) \in \RR^{n}\), we may take an \(\alpha\)-weighted sum
over \cref{eq:conj-cov}. By \cref{lem:score}, \(E[u] = 0\), thus
\cref{eq:conj-cov} becomes
\[
E_{\H\sim\rho}\Big[u(H) \dv{}{t} \log \rho(\H)\Big] = -\Cov_{\H \sim \rho} \Big[u(H), \L(\H)\Big].
\]
This implies that the second term (first term on the right)  in
\cref{eq:price-chain-log} and the second term (first term on the right) in
\cref{eq:price-thm}  are equivalent, and, therefore, \cref{eq:price-chain-log}
implies
\cref{eq:price-thm}.
\end{Proof}

\subsection{Proof: Continuous Inference} \label{sec:prf-bayes}

In this section, we give a derivation of the discrete-time replicator equation
as background. We reference this derivation while also proving that continuous
Bayesian inference (\cref{eq:bayes}) may be used to derive Bayes's rule
(\cref{thm:bayes}) and that \ngd provides an optimal approximation of continuous
Bayesian inference (\cref{thm:oci}).

Let us first establish that the replicator dynamics preserve the normalization
condition necessary for a proper probability distribution, restating the
replicator equation (\cref{eq:replicator}) for local reference:
\begin{equation*}
  \dot{\rho}(\h) = \rho(\h) \Big[ \avgL_{\rho} - \L(\h) \Big],
  \quad\text{where}\quad
  \avgL_{\rho} \define \sum_{\h} \rho(\h) \L(\h), \quad \sum_{\h} \rho(\h) = 1. \tag{\ref*{eq:replicator}}
\end{equation*}

\begin{lem}
  \label{lem:normalization}
  \emph{(Preservation of Normalization)}.
  The dynamics of the continuous time replicator equation preserve the
normalization of \(\rho\) (\ie, \(\sum_{\h} \rho(\h) = 1\)).  That is,
\(\dv{}{t} \sum_{\h} \rho(\h) = 0\).
\end{lem}

\begin{proof}
\[
\dv{}{t} \sum_{\h} \rho(\h)
= \sum_{\h} \dot{\rho}(\h)
= \sum_{\h} \rho(\h) \left[ \avgL_{\rho} - \L(\h) \right]
= \avgL_{\rho} - \sum_{\h} \rho(\h) \L(\h)
= 0.
\]
\end{proof}

The replicator equation is frequently encountered in discrete time.
\begin{lem}
  \label{lem:replicator-discrete}
  \emph{(The Discrete-Time Replicator Equation)}.
Define
\[
  \log r_{t}(\h) = -\frac{1}{\Delta t}\int_{t}^{t+\Delta t} \L_{t'}(\h) \dd{t'},
\]
for each \(h\), as the time-average of \(-\L_{t'}(\h)\) over
\([t, t + \Delta t)\), and let
\[
\wtilde{\r}_{t}(\Delta t) \define \sum_{\h} \rho_{t}(\h) \r_{t}(\h)^{\Delta t}.
\]
It follows that
\begin{equation} \label{eq:replicator-discrete}
  \rho_{(t+\Delta t)}(\h) = \rho_{t}(\h) \frac{\r_{t}(\h)^{\Delta t}}{\wtilde{\r}_{t}(\Delta t)}.
\end{equation}
\end{lem}

\begin{proof}
The solution of \cref{eq:replicator} (which may be verified by differentiating
with respect to time) is
\begin{equation*}
    \rho_{(t + \Delta t)}(\h)
       = \rho_{t}(\h)
       \Big(\underbrace{ \exp \int_{t}^{(t + \Delta t)} \avgL_{\rho_{t'}} \dd{t'}}_{C_{t, \Delta t}} \Big)
       \Big(\underbrace{\exp \int_{t}^{(t + \Delta t)} - \L_{t'}(\h) \dd{t'}}_{\r_{t}(\h)^{\Delta t}}\Big).
  \end{equation*}
  After summing over \(\h\) on both sides of this equation, normalization
(\cref{lem:normalization}) implies that the constant \(C_{t, \Delta t}\) is
necessarily equal to the multiplicative inverse of
\(\wtilde{\r}_{t}(\Delta t) \define \sum_{\h} \rho_{t}(\h) \r_{t}(\h)^{\Delta t}\).
\end{proof}

Our formulation of ``continuous Bayesian inference'' in the main text defines,
for local reference,
\begin{equation*}
  \h(x_{t} ; t) = \Pr_{\h}(\X_{t}{=}\x_{t} \smid t)
  \quad ; \quad
  \n(x_{t} ; t) = \Pr_{\n}(\X_{t}{=}\x_{t} \smid t)
  \quad ; \quad
  \L(\h, t) = -\log \h(\x_{t} ; t).
  \tag{\ref*{eq:surprisal}}
\end{equation*}
such that
\begin{equation*}
  \dot{\rho}_{t}(\h, \x_{t}) = \rho_{t}(\h) \Big[ \avgL_{\rho_{t}}(x_{t}) + \log \h(\x_{t} ; t) \Big],
  \quad \text{where} \quad
  \avgL_{\rho_{t}}(x_{t}) = -\sum_{\h} \rho_{t}(\h) \log \h(\x_{t} ; t).
  \tag{\ref*{eq:bayes}}
\end{equation*}

\thmbayes*

\begin{Proof}{of \cref*{thm:bayes}}\label{prf:bayes}
To discretize \cref{eq:bayes}, we first denote the \emph{path} of observations
over the time interval from \(t\) up to \(t + \Delta t\) as
\(x_{t}^{\Delta t} \define \{x_{t'} \colon t' \in [t, t + \Delta t)\}\). Next, define
the probability density of the path to be proportional to the product of the
probabilities of its instantaneous values.
\begin{equation}
  \log \h(\x_{t}^{\Delta t} ; t) \define \frac{1}{[\mathsf{t}]} \int_{t}^{t + \Delta t} \log \h(x_{t'} ; t') \dd{t'}.
\end{equation}
Note that we normalize this equation to make it properly dimensionless by choice
of an arbitrary scale, where \([\mathsf{t}]\) denotes units of time. The choice
of an arbitrary scale is integral to the definition of differential entropy, as
it allows us to establish a volume of configuration space (in this case, with
units of time) to correspond to unit entropy. For a motivating example, we must choose how
many units of (differential) entropy correspond to the space of possible paths
over 1\(\mathsf{s}\), when each \(X_{t}\) is a uniformly distributed Bernoulli
random variable. We choose the same units that we use to measure \(\Delta t\),
so that \([\mathsf{t}]\) may be considered equal to 1 hereafter.

Subject to the loss of \cref{eq:surprisal}, when
\(\L(\h, t) = - \log \h(X_{t}^{\Delta t}, t)\), retracing the derivation
of the discrete-time replicator equation (\cref{lem:replicator-discrete}) yields
Bayes's rule, \ie,
\begin{equation} \label{eq:bayes-discrete}
  \rho_{(t+\Delta t)}(\h \smid \X_{t}^{\Delta t})
  = \rho_{t}(\h) \frac{\h(X_{t}^{\Delta t} , t)}{\Pr_{\rho_{t}}(X_{t}^{\Delta t})},
  \quad \text{from} \quad
  \rho_{(t+\Delta t)}(\h)
  = \rho_{t}(\h) \frac{\r_{t}(\h)^{\Delta t}}{\wtilde{\r}_{t}(\Delta t)},
\end{equation}
where
\[
\r_{t}(\h)^{\Delta t} = \exp \int_{t}^{(t + \Delta t)} - \L_{t'}(\h) \dd{t'} = \h(X_{t}^{\Delta t} , t)
\]
and
\[
\wtilde{\r}_{t}(\Delta t) \define \sum_{\h} \rho_{t}(\h) \r_{t}(\h)^{\Delta t}  = \Pr_{\rho_{t}}(\X_{t}^{\Delta t}).
\]
We identify
\(
\rho_{t + \Delta(t)}(\h)
\)
as the posterior
\(
\rho_{t + \Delta(t)}(\h \smid \X_{t}^{\Delta t})
\)
when path \(X_{t}^{\Delta t}\) is observed.
\end{Proof}

Having established that the replicator equation with a specific loss based on
\emph{surprisal} (\cref{eq:surprisal}) may be identified with ``continuous
Bayesian inference'', we next prove that continuous Bayesian inference is
optimally approximated by \ngd (\cref{thm:oci}). Let us first restate the
definition of Kullback-Leibler divergence of \(\n\) from \(\h\), to which we
relate the gradient of the surprisal-based loss:
\begin{equation*}
  \DD_{t}(\n \parallel \h) \define -\sum_{x} \n(\x ; t) \log \frac{\h(\x ; t)}{\n(\x ; t)}.
  \tag{\ref*{eq:kl}}
\end{equation*}

\lemgradients*

\begin{Proof}{of \cref*{lem:gradients}}\label{prf:gradients}
  \begin{align*}
    & \E_{\H\sim\rho_{t}}[\DD(\H, t)] - \E_{\X_{t}\sim\n}[\avgL_{\rho_{t}}(X_{t})] \\
    &\qquad =
      -\sum_{\x, \h} \rho_{t}(\h) \n(\x ; t) \log \frac{\h(\x ; t)}{\n(\x ; t)}
      +\sum_{\x, \h} \rho_{t}(\h) \n(\x ; t) \log \h(\x_{t} ; t) \\
    &\qquad =
      \sum_{\x,\h} \rho_{t}(\h)\n(\x, t) \log \n(\x, t) \\
    &\qquad =
  \sum_{\x} \n(\x, t) \log \n(\x, t).
  \end{align*}
  As the negative entropy of \(X_{t}\sim\n\), this difference is independent of
  \(\rho_{t}\) and therefore has zero gradient with respect to \(\rho_{t}\). Each
  term of the original expression must therefore have the same gradient.
\end{Proof}

We conclude with a restatement and proof of the optimal correspondence of \ngd
with the appropriate loss to continuous Bayesian inference:

\thmoci*

\begin{Proof}{of \cref*{thm:oci}}\label{prf:oci}
Assume a twice-differentiable parameterization for probability distribution
\(\rho(\h ; \theta)\).  \ngd of \(\E_{\H \sim \rho}[\DD(\H, t)]\) (\cref{eq:kl})
is the same as \ngd of \(\E_{X_{t} \sim \n}[\avgL_{\rho_{t}}(X_{t})]\)
(\cref{eq:bayes}), since these quantities have equivalent gradients
(\cref{lem:gradients}). We will treat the latter.

By \cref{thm:cns}, \ngd of \(\E_{X_{t} \sim \n}[\avgL_{\rho_{t}}(X_{t})]\) with
respect to \(\theta\) provides an optimal approximation of the replicator
dynamics (\cref{eq:replicator}) with the corresponding stochastic loss
\(\L(\h) = -\log \h(x_{t} ; t)\).

Finally, because \cref{thm:bayes} indicates that the replicator dynamics
are consistent with continuous time Bayesian inference, it follows that
\ngd of \(\E_{H \sim \rho_{t}}[\DD(H), t]\) with respect to \(\theta\) is
an optimal approximation of Bayesian inference in continuous time.
\end{Proof}

%% file: main.bbl
\begin{thebibliography}{28}
\providecommand{\natexlab}[1]{#1}
\providecommand{\url}[1]{\texttt{#1}}
\expandafter\ifx\csname urlstyle\endcsname\relax
  \providecommand{\doi}[1]{doi: #1}\else
  \providecommand{\doi}{doi: \begingroup \urlstyle{rm}\Url}\fi

\bibitem[Amari(1998)]{amari1998natural}
Shun-Ichi Amari.
\newblock Natural gradient works efficiently in learning.
\newblock \emph{Neural computation}, 10\penalty0 (2):\penalty0 251--276, 1998.

\bibitem[B{\"a}ck et~al.(2018)B{\"a}ck, Fogel, and
  Michalewicz]{back2018evolutionary}
Thomas B{\"a}ck, David~B Fogel, and Zbigniew Michalewicz.
\newblock \emph{Evolutionary computation 1: Basic algorithms and operators}.
\newblock CRC press, 2018.

\bibitem[Bloembergen et~al.(2015)Bloembergen, Tuyls, Hennes, and
  Kaisers]{bloembergen2015evolutionary}
Daan Bloembergen, Karl Tuyls, Daniel Hennes, and Michael Kaisers.
\newblock Evolutionary dynamics of multi-agent learning: A survey.
\newblock \emph{Journal of Artificial Intelligence Research}, 53:\penalty0
  659--697, 2015.

\bibitem[Chalub et~al.(2021)Chalub, Monsaingeon, Ribeiro, and
  Souza]{chalub2021gradient}
Fabio~ACC Chalub, L{\'e}onard Monsaingeon, Ana~Margarida Ribeiro, and Max~O
  Souza.
\newblock Gradient flow formulations of discrete and continuous evolutionary
  models: a unifying perspective.
\newblock \emph{Acta Applicandae Mathematicae}, 171\penalty0 (1):\penalty0
  1--50, 2021.

\bibitem[Cressman and Tao(2014)]{cressman2014replicator}
Ross Cressman and Yi~Tao.
\newblock The replicator equation and other game dynamics.
\newblock \emph{Proceedings of the National Academy of Sciences}, 111\penalty0
  (supplement\_3):\penalty0 10810--10817, 2014.

\bibitem[Fradkov(2020)]{fradkov2020early}
Alexander~L Fradkov.
\newblock Early history of machine learning.
\newblock \emph{IFAC-PapersOnLine}, 53\penalty0 (2):\penalty0 1385--1390, 2020.

\bibitem[Friedman(1991)]{friedman1991evolutionary}
Daniel Friedman.
\newblock Evolutionary games in economics.
\newblock \emph{Econometrica: journal of the econometric society}, pages
  637--666, 1991.

\bibitem[Friedman and Sinervo(2016)]{friedman2016evolutionary}
Daniel Friedman and Barry Sinervo.
\newblock \emph{Evolutionary games in natural, social, and virtual worlds}.
\newblock Oxford University Press, 2016.

\bibitem[Gao and Pavel(2017)]{gao2017properties}
Bolin Gao and Lacra Pavel.
\newblock On the properties of the softmax function with application in game
  theory and reinforcement learning.
\newblock \emph{arXiv preprint arXiv:1704.00805}, 2017.

\bibitem[Harper(2009{\natexlab{a}})]{harper2009information}
Marc Harper.
\newblock Information geometry and evolutionary game theory.
\newblock \emph{arXiv preprint arXiv:0911.1383}, 2009{\natexlab{a}}.

\bibitem[Harper(2009{\natexlab{b}})]{harper2009replicator}
Marc Harper.
\newblock The replicator equation as an inference dynamic.
\newblock \emph{arXiv preprint arXiv:0911.1763}, 2009{\natexlab{b}}.

\bibitem[Harper(2011)]{harper2011escort}
Marc Harper.
\newblock Escort evolutionary game theory.
\newblock \emph{Physica D: Nonlinear Phenomena}, 240\penalty0 (18):\penalty0
  1411--1415, 2011.

\bibitem[Harper and Safyan(2020)]{harper2020momentum}
Marc Harper and Joshua Safyan.
\newblock Momentum accelerates evolutionary dynamics.
\newblock \emph{arXiv preprint arXiv:2007.02449}, 2020.

\bibitem[Hennes et~al.(2019)Hennes, Morrill, Omidshafiei, Munos, Perolat,
  Lanctot, Gruslys, Lespiau, Parmas, Duenez-Guzman, et~al.]{hennes2019neural}
Daniel Hennes, Dustin Morrill, Shayegan Omidshafiei, Remi Munos, Julien
  Perolat, Marc Lanctot, Audrunas Gruslys, Jean-Baptiste Lespiau, Paavo Parmas,
  Edgar Duenez-Guzman, et~al.
\newblock Neural replicator dynamics.
\newblock \emph{arXiv preprint arXiv:1906.00190}, 2019.

\bibitem[Hofbauer et~al.(1998)Hofbauer, Sigmund,
  et~al.]{hofbauer1998evolutionary}
Josef Hofbauer, Karl Sigmund, et~al.
\newblock \emph{Evolutionary games and population dynamics}.
\newblock Cambridge university press, 1998.

\bibitem[Hu et~al.(2022)Hu, Ao, So, Yang, and Wen]{hu2022riemannian}
Jiang Hu, Ruicheng Ao, Anthony Man-Cho So, Minghan Yang, and Zaiwen Wen.
\newblock Riemannian natural gradient methods.
\newblock \emph{arXiv preprint arXiv:2207.07287}, 2022.

\bibitem[Khan and Rue(2021)]{khan2021bayesian}
Mohammad~Emtiyaz Khan and H{\aa}vard Rue.
\newblock The {B}ayesian learning rule.
\newblock \emph{arXiv preprint arXiv:2107.04562}, 2021.

\bibitem[Littlestone and Warmuth(1994)]{littlestone1994weighted}
Nick Littlestone and Manfred~K Warmuth.
\newblock The weighted majority algorithm.
\newblock \emph{Information and computation}, 108\penalty0 (2):\penalty0
  212--261, 1994.

\bibitem[Lloyd(2020)]{sep-selection-units}
Elisabeth Lloyd.
\newblock Units and levels of selection.
\newblock In Edward~N. Zalta, editor, \emph{The {Stanford} Encyclopedia of
  Philosophy}. Metaphysics Research Lab, Stanford University, {S}pring 2020
  edition, 2020.

\bibitem[Martens(2020)]{martens2020new}
James Martens.
\newblock New insights and perspectives on the natural gradient method.
\newblock \emph{The Journal of Machine Learning Research}, 21\penalty0
  (1):\penalty0 5776--5851, 2020.

\bibitem[M{\"u}ller and Mont{\'u}far(2022)]{muller2022geometry}
Johannes M{\"u}ller and Guido Mont{\'u}far.
\newblock Geometry and convergence of natural policy gradient methods.
\newblock \emph{arXiv preprint arXiv:2211.02105}, 2022.

\bibitem[Nurbekyan et~al.(2022)Nurbekyan, Lei, and
  Yang]{nurbekyan2022efficient}
Levon Nurbekyan, Wanzhou Lei, and Yunan Yang.
\newblock Efficient natural gradient descent methods for large-scale
  optimization problems.
\newblock \emph{arXiv preprint arXiv:2202.06236}, 2022.

\bibitem[Otwinowski et~al.(2020)Otwinowski, LaMont, and
  Nourmohammad]{otwinowski2020information}
Jakub Otwinowski, Colin~H LaMont, and Armita Nourmohammad.
\newblock Information-geometric optimization with natural selection.
\newblock \emph{Entropy}, 22\penalty0 (9):\penalty0 967, 2020.

\bibitem[Peirson et~al.(2022)Peirson, Amid, Chen, Feinberg, Warmuth, and
  Anil]{peirson2022fishy}
Abel Peirson, Ehsan Amid, Yatong Chen, Vladimir Feinberg, Manfred~K Warmuth,
  and Rohan Anil.
\newblock Fishy: Layerwise fisher approximation for higher-order neural network
  optimization.
\newblock In \emph{Has it Trained Yet? NeurIPS 2022 Workshop}, 2022.

\bibitem[Queller(2017)]{queller2017fundamental}
David~C Queller.
\newblock Fundamental theorems of evolution.
\newblock \emph{The American Naturalist}, 189\penalty0 (4):\penalty0 345--353,
  2017.

\bibitem[Sandholm(2010)]{sandholm2010population}
William~H Sandholm.
\newblock \emph{Population games and evolutionary dynamics}.
\newblock MIT press, 2010.

\bibitem[Sinervo and Calsbeek(2006)]{sinervo2006developmental}
Barry Sinervo and Ryan Calsbeek.
\newblock The developmental, physiological, neural, and genetical causes and
  consequences of frequency-dependent selection in the wild.
\newblock \emph{Annu. Rev. Ecol. Evol. Syst.}, 37:\penalty0 581--610, 2006.

\bibitem[Tappert(2020)]{tappertfrank}
Charles~C Tappert.
\newblock Frank {R}osenblatt, the father of deep learning.
\newblock \emph{Proceedings of Student-Faculty Research Day, CSIS, Pace
  University}, 2020.

\end{thebibliography}
